%% file: main.tex
\documentclass[10pt,twocolumn,letterpaper]{article}

 \usepackage{cvpr}              
\usepackage{microtype}

\input{preamble}

\definecolor{cvprblue}{rgb}{0.21,0.49,0.74}
\usepackage[pagebackref,breaklinks,colorlinks,citecolor=cvprblue]{hyperref}

\newenvironment{packed_lefty_item}{
\begin{itemize}[leftmargin=*]
\vspace{-0pt}
  \setlength{\itemsep}{0pt}
  \setlength{\parskip}{0pt}
  \setlength{\parsep}{0pt}
  \setlength{\topsep}{-0pt}
  \setlength{\partopsep}{0pt}
}{\end{itemize}\vspace{-0pt}}

\usepackage[capitalize]{cleveref}
\crefname{section}{Sec.}{Secs.}
\Crefname{section}{Section}{Sections}
\Crefname{table}{Table}{Tables}
\crefname{table}{Tab.}{Tabs.}

\newcommand{\Ours}{LidaRF}
\newcommand{\Ourstight}{LidaRF}

\def\paperID{336} 
\def\confName{CVPR}
\def\confYear{2024}

\title{\Ourstight: Delving into Lidar for Neural Radiance Field on Street Scenes}

\author{Shanlin Sun$^1$ \quad Bingbing Zhuang$^3$ \quad Ziyu Jiang$^3$ \quad Buyu Liu$^3$ \\
Xiaohui Xie$^1$ \quad Manmohan Chandraker$^{2,3}$  \vspace{+0.3em} \\
$^1$University of California, Irvine~~~$^2$University of California, San Diego~~~$^3$NEC Labs America\vspace{-0em} \\
}

\begin{document}
\maketitle
\input{sec/0_abstract}    
\input{sec/1_intro}

\input{sec/2_related}

\input{sec/3_method}

\input{sec/4_experiments}
\input{sec/6_conclusion}
{
    \small
    \bibliographystyle{ieeenat_fullname}
    \bibliography{main}
}

\clearpage
\maketitlesupplementary
\input{sec/X_suppl}

\end{document}

%% file: preamble.tex
\usepackage[dvipsnames]{xcolor}
\usepackage{multirow}
\usepackage{graphicx}
\usepackage{arydshln}
\usepackage{lipsum} 
\usepackage{float}

\newcommand{\tickbox}{{\makebox[0pt][l]{$\square$}\raisebox{.15ex}{\hspace{0.1em}$\checkmark$}}}
\newcommand{\untickbox}{{\makebox[0pt][l]{$\square$}}}


%% file: sec/0_abstract.tex
\begin{abstract}

Photorealistic simulation plays a crucial role in applications such as autonomous driving, where advances in neural radiance fields (NeRFs) may allow better scalability through the automatic creation of digital 3D assets. However, reconstruction quality suffers on street scenes due to largely collinear camera motions and sparser samplings at higher speeds. On the other hand, the application often demands rendering from camera views that deviate from the inputs to accurately simulate behaviors like lane changes. In this paper, we propose several insights that allow a better utilization of Lidar data to improve NeRF quality on street scenes. First, our framework learns a geometric scene representation from Lidar, which are fused with the implicit grid-based representation for radiance decoding, thereby supplying stronger geometric information offered by explicit point cloud. Second, we put forth a robust occlusion-aware depth supervision scheme, which allows utilizing densified Lidar points by accumulation. Third, we generate augmented training views from Lidar points for further improvement. Our insights translate to largely improved novel view synthesis under real driving scenes.
\end{abstract}

%% file: sec/1_intro.tex
\section{Introduction}
\label{sec:intro}

\begin{figure}[t!]
  \centering  \includegraphics[width=1.0\linewidth, trim = 8mm 25mm 100mm 0mm, clip]{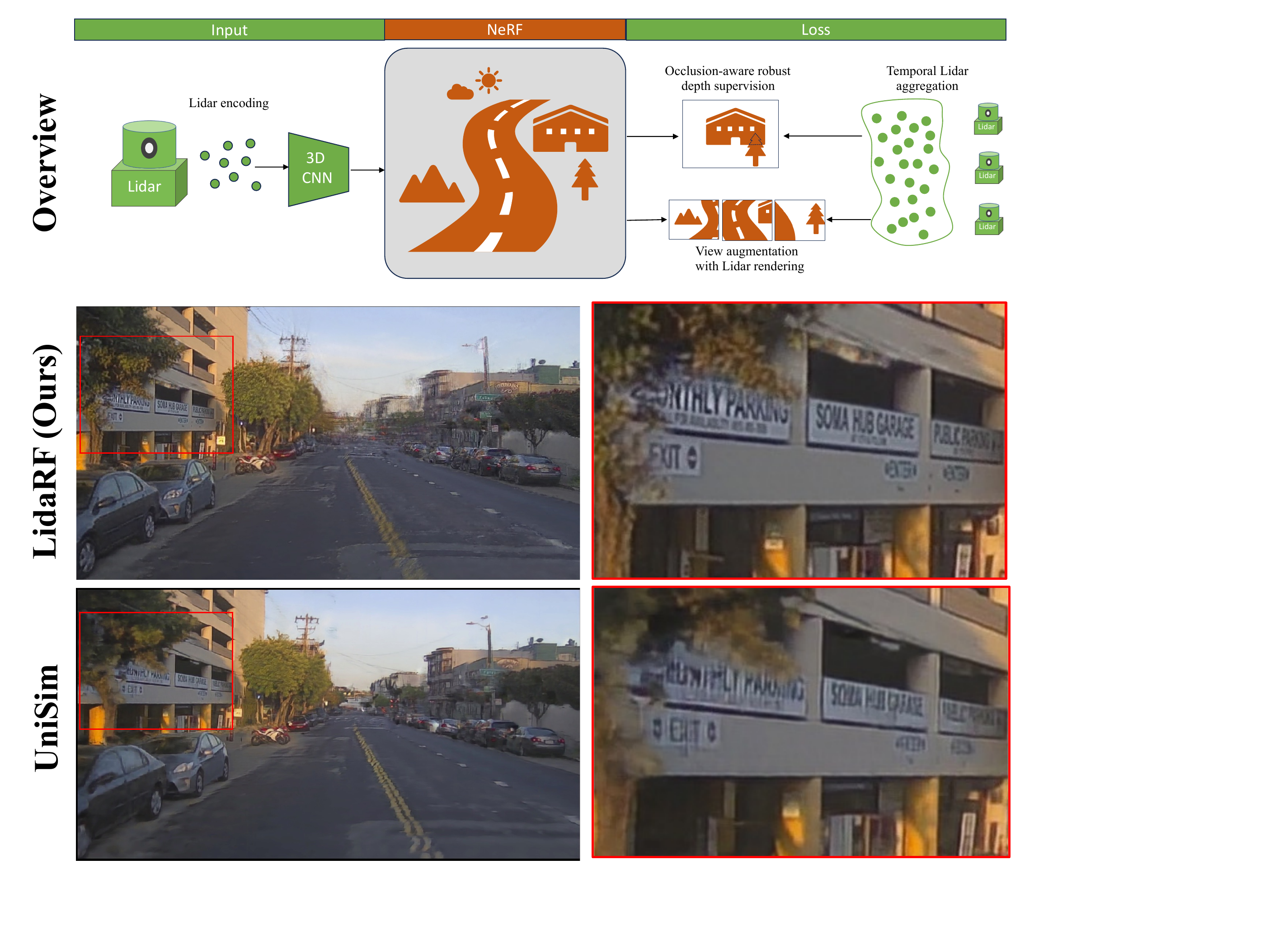}
  \centering
  \vspace{-2em}
  \caption{Our framework leverages Lidar to a deep extent to unlock its potential for neural rendering on street scenes, leading to state-of-the-art performance in comparison to UniSim~\cite{yang2023unisim}.}
  \label{fig:teaser} 
  \vspace{-1em}
\end{figure}

Photorealistic simulation is needed in many applications like autonomous driving, where it is hard to ensure the diversity and coverage of real data. More critically, extensive verification has to be carried out in simulated environment before road testing, in order to ensure safety.
Traditional simulation pipeline typically requires graphic artists to manually create 3D assets and compose into virtual environment of interest. However, the demand on human efforts and expertise has prevented it from being scalable in practice. 


Neural Radiance Field (NeRF~\cite{mildenhall2020nerf}) has recently emerged as a promising way to approach simulation.
NeRF has proven an effective implicit representation of scene radiance, with remarkable abilities to capture and interpolate appearances. However, such good performance often requires dense view coverage in training data, in order for sufficient constraints to learn accurately the underlying geometry, material properties and illumination.
While dense view coverage is not much of a problem in controlled environments, it poses challenges on street scenes -- the data collection vehicle typically drives forward along lanes at a potentially high speed, leading to sparse and nearly collinear camera paths. Besides sparsity, forward camera trajectories are well-known~\cite{vedaldi2007moving} to be challenging for 3D reconstruction as it provides much weaker multi-view geometric constraints. Furthermore, road surface is typically low-texture, further introducing ambiguity in scene reconstruction.   





To address these challenges, and in view of the implicit representation of NeRF lacking explicit geometric constraint, our key idea lies in leveraging Lidar as an explicit complementary to NeRF. 
Despite Lidar being used in existing works~\cite{xie2022s,rematas2022urban,yang2023unisim} for street scenes, high-quality rendering remains challenging. Our work delves into Lidar (dubbed \Ourstight) and reaps its benefits to a greater extent, yielding largely improved view synthesis quality as shown in \cref{fig:teaser}.

Our contributions are three-fold. (i) \textit{Fusing Lidar encoding and grid feature for enhanced scene representation}. 
While Lidar has been applied as a natural depth supervision, involving Lidar in the NeRF input offers great potential for geometric inductive bias but remains less straightforward. To this end, we adopt the grid-based representation~\cite{mueller2022instant} but fuse features learned from point clouds into the grid, to inherit benefits from the explicit point cloud representation. Inspired by the success of 3D perception frameworks~\cite{yin2021center,liu2023bevfusion}, we leverage 3D sparse convolution network as an effective and efficient architecture to extract geometric features from the local and global context of Lidar point clouds.
(ii) \textit{Robust occlusion-aware depth supervision.}
Similar to existing works~\cite{yang2023unisim, xie2022s}, we also apply Lidar as a source of depth supervision, but do so to a greater degree. Due to the sparsity of Lidar points limiting its utility, especially in low-texture regions, we densify Lidar points across nearby frames to generate denser depth maps. 
However, the depth map so obtained does not account for occlusion, yielding ghost depth supervision. 
Hence, we put forth a robust depth supervision scheme in a curriculum learning fashion -- supervising depth from near to far field while gradually filtering out bogus depth as the NeRF trains, leading to more effective learning of depth from Lidar.   
(iii) \textit{Lidar-based view augmentation.} Furthermore, in view of the view sparsity and limited coverage in the driving scene,  we leverage Lidar to denstify training views. That is, we project accumulated Lidar points to novel training views; note they could be views deviating from the driving trajectories to some extent. 
These views projected from Lidar are added into the pool of training data, but recall that they do not account for occlusion. However, we apply the fore-mentioned supervision scheme to address the occlusion issue, yielding improved performance.


While our insights are also applicable for general scenes, we focus the evaluation on street scenes in this work, which leads to significant improvement compared to prior art, both quantitatively and qualitatively. 
Our \Ours also shows advantage in interesting applications such as lane changes that require greater deviation from input views.

In summary, our proposed insights on better incorporation of Lidar lead to significantly improved quality of NeRF in challenging street scene applications.

%% file: sec/2_related.tex
\section{Related Work}
\label{sec:related_work}

\noindent \textbf{NeRF fundamentals.} 
Neural radiance field (NeRF)~\cite{mildenhall2020nerf} has become a widely applied scene representation thanks to its remarkable capability for photorealistic
novel view synthesis. Since its advent, rapid progress has been made to increase its range of applicability. Notably, given the positional encoding in NeRF does not account for scene scale and causes aliasing when training and testing images are of different resolutions, Mip-NeRF~\cite{barron2021mip} proposes to anti-alias it by casting conical frustums instead of rays, with integrated positional encoding with scale awareness. Mip-NeRF 360~\cite{barron2022mip} further extends it to handle unbounded scenes by a nonlinear mapping of space for scene contraction. In view of querying large MLPs being the bottleneck for efficiency, Instant-NPG~\cite{mueller2022instant} proposes the grid-based scene representation stored using a hash map, with the learned features decoded into radiance by a tiny and fast MLP~\cite{tiny-cuda-nn}. We build our framework on top of the successful recipes of Mip-NeRF 360 and Instant-NPG, but makes important advancement to seamlessly integrate the valuable information offered by additional Lidar sensors.  

\noindent \textbf{Point-based NeRFs.} 
In contrast to the implicit scene representation, point could is an explicit representation holding the advantage of capturing accurate scene geometry. Point cloud is also widely available, either from structure-from-motion and multi-view stereo, or directly from time-of-flight depth sensors like Kinect or Lidar. This leads to the line of research~\cite{aliev2020neural, yang2023nerfvs, ost2022neural,zhang2023frequency} in rendering images from point clouds or surfaces. Point-NeRF~\cite{xu2022point} represents one of the pioneering works that assign image features to the point clouds, from which the radiance filed along the rendering ray is decoded by querying features from nearby points. Point2Pix~\cite{hu2023point2pix} utilizes point encoding to render point clouds in indoor scenes to images. TriVol~\cite{hu2023trivol} adopts triple slim volume to encode point cloud efficiently. Pointersect~\cite{chang2023pointersect} proposes to render point clouds by directly inferring the intersection of the ray with the underlying surface. However, these works only demonstrate results on objects or small-scale indoor scenes. 
Chang et al.~\cite{chang2023neural} has recently extended Point-NeRF to street scenes, but the rendering remains low-resolution and incomplete due to the sparse nature of Lidar point clouds. In contrast, our \Ours fuses Lidar encoding with high-resolution grid-based representation for feature learning, yielding results far superior to \cite{chang2023neural}.

\noindent \textbf{Street scene NeRFs.} The need for photorealistic simulation in autonomous driving inspires researches~\cite{ost2021neural, kundu2022panoptic, tancik2022block, xie2022s, zhang2023nerflets, yang2023unisim,  liu2023real, wang2023neural,guo2023streetsurf,wimbauer2023behind} to explore NeRFs on unconstrained street scenes. Notably, UniSim~\cite{yang2023unisim} demonstrates the promising applicability of NeRF for closed-loop simulation of the autonomy, taking only a real driving log as input.
Some works focus on handling sparse view observations~\cite{zhou2023sampling, carlson2023cloner} or improving geometry of NeRF~\cite{wang2023planerf, guo2023streetsurf}.
Despite these efforts, high-quality rendering of street scenes remains challenging for NeRF. 
Our work improves NeRF by leveraging Lidar data to a greater extent, with insights on hybrid feature encoding and robust depth supervision with densified Lidar. It is worth noting that while S-NeRF~\cite{xie2022s} also densifies Lidar with a depth completion network, our strategy distinguishes itself by relying only on actual Lidar frames without additional training data, and is not affected by potential errors in the depth prediction.     

\noindent \textbf{Depth-supervised NeRFs.} The vanilla NeRF does not enforce any explicit constraint on geometry, often leading to inaccurate depth or surface recovery. This inspires many works to impose depth supervision in various forms~\cite{kangle2021dsnerf,wang2023sparsenerf,yang2023nerfvs,wang2023digging,wei2023depth,yu2022monosdf}. Notably, NeRFs on street scenes~\cite{xie2022s,yang2023unisim,rematas2022urban} typically involve derived depth supervision from Lidar. Our work distinguishes itself by reap benefits from Lidar to a deeper extent, by Lidar aggregation while accounting for occlusions.


\section{Preliminaries}
\label{sec:preliminary}

\noindent \textbf{Neural Radiance Field (NeRF)}~\cite{mildenhall2020nerf} represents a radiance field with a continuous neural network $f: (\mathbf{x}, \mathbf{d}) \rightarrow (c, \sigma)$, mapping spatial location $\mathbf{x} = (x,y,z)$ and viewing direction $\mathbf{d} = (\theta, \phi)$ to the RGB color $c$ and volumetric density $\sigma$ at that point. 
The network is queried at each point along the rendering ray to estimate color and density, which are then composed into the final pixel color using the volume rendering equation~\cite{kajiya1984ray}.
NeRF is optimized through the loss function $\mathcal{L}_{\text{rgb}}$ defined as the mean squared error between the predicted and true colors of the training RGB images.

\noindent \textbf{Nerfacto} is the recommended approach within the open-source project Nerfstudio~\cite{nerfstudio}, integrating a variety of recipes that have proven effective for a wide range of real-world data. First, Nerfactor is capable of handling unbound scenes as required by street scenes, by applying the scene contraction strategy as in MipNeRF-360~\cite{barron2022mip}. For efficiency, it follows~\cite{barron2022mip} to use proposal networks with small MLPs to consolidate the sampled locations along each ray to the regions near the first surface intersection. 
In terms of NeRF network architecture, it follows Instant-NGP~\cite{mueller2022instant} to leverage the grid-based feature representation parameterized by a hash map, which allows to decode color and density with the so-called fused MLPs, \ie small MLPs that admits fast implementation ~\cite{tiny-cuda-nn}. 



%% file: sec/3_method.tex
\section{Method}
\label{sec:method}

\input{figure/pipeline}

\input{sec/3_1_overview}
\input{sec/3_2_lenc}
\input{sec/3_3_ds}
\input{sec/3_5_aug}

%% file: figure/pipeline.tex
\begin{figure*}
  \centering
  \includegraphics[width=0.9\linewidth]{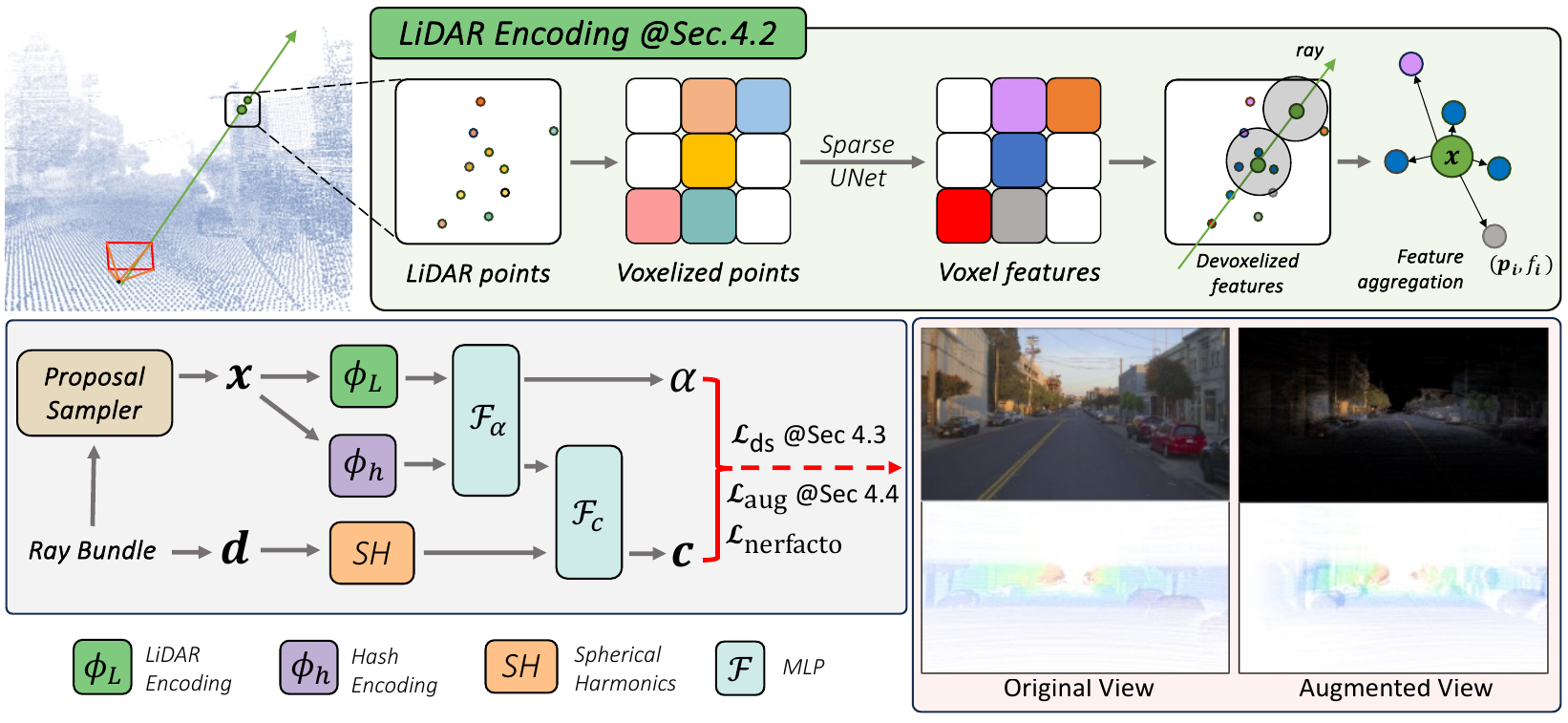}
  \centering
  \caption{
  \textbf{Overview of \Ourstight} -- it takes as input the sampled 3D positions $\textbf{x}$ and ray directions $\textbf{d}$, and outputs corresponding density $\alpha$ and color $\textbf{c}$. It incorporates both hash encoding and LiDAR encoding using a sparse UNet. Additionally, augmented training data is generated through LiDAR projections, and the geometry prediction is trained with our proposed robust depth supervision scheme.
  }
  \label{fig:pipeline} 
  \vspace{-1em}
\end{figure*}

%% file: sec/3_1_overview.tex
\subsection{Overview}
\label{subsec:overview}


We build \Ours on top of Nerfacto, but develop several insights to integrate the use of Lidar for high-quality view synthesis on street scenes, as illustrated in \cref{fig:pipeline}. Our pipeline takes Lidar point clouds as input (\cref{subsec:Lidar_enc}), extracts geometric features from the point clouds, and fuses it with the hash-based feature grid to combine their complementary benefits. The hybrid features are fed to MLPs for decoding color and density, followed by standard volume rendering.
In the output, we leverage denser Lidar points accumulated across frames as depth supervision while accounting for occlusion. This leads to two extra losses $\mathcal{L}_{\text {ds}}$ (\cref{subsec:ds}) and $\mathcal{L}_{\text {aug}}$ (\cref{subsec:avs}), besides the original losses from Nerfactor $\mathcal{L}_{\text {nerfacto}}$. All together, our loss $\mathcal{L}$ is written as

\vspace{-1em}
\begin{align}
\mathcal{L} = 
\mathcal{L}_{\text {nerfacto}} + 
\lambda_1 \cdot \underbrace{\mathcal{L}_{\text {ds}}}_{\text {Sec. } 4.3} +
\lambda_2 \cdot \underbrace{\mathcal{L}_{\text {aug}}}_{\text {Sec. } 4.4}, \\
\text{where}~~\mathcal{L}_{\text {nerfacto}} = \mathcal{L}_{\text {rgb}} + \lambda_3 \cdot \mathcal{L}_{\text {dist}} + \lambda_4 
\cdot \mathcal{L}_{\text {interval}}. 
\end{align}

%% file: sec/3_2_lenc.tex
\subsection{Hybrid Representation with Lidar Encoding}
\label{subsec:Lidar_enc}

\noindent \textbf{Motivation.} Lidar point clouds holds strong potential for geometric guidance that is highly valuable for NeRF. However, relying on Lidar features alone for scene representation, as done in~\cite{chang2023neural}, results in low-resolution rendering, due to the sparse nature of Lidar points despite temporal accumulation. In addition, Lidar does not cover the entire scene due to its limited field of view, e.g. it does not capture building surface above a certain height, yielding blank rendering in those regions as in~\cite{chang2023neural}. Our framework, in contrast, fuses the Lidar features with the high-resolution spatial grid of features to leverage the strength of both, which are jointly learned for high-quality and complete scene rendering.

\vspace{0.1cm}
\noindent \textbf{Lidar feature extraction.} 
We detail here the extraction of geometric features for each Lidar point.
Referring to \cref{fig:pipeline}, we first aggregate Lidar point clouds from all frames of the entire sequence to construct a denser set of point clouds. We then voxelize the point clouds into a voxel grid, where the spatial position of points falling into each voxel cell are averaged, yielding a 3-dim feature for each voxel cell. 
Inspired by its wide success on 3D perception frameworks~\cite{yin2021center,liu2023bevfusion}, 
we encode the scene geometry feature with a 3D sparse UNet~\cite{choy20194d,tangandyang2023torchsparse} on the voxel grid, which permits learning from a more global context of the scene geometry. The 3D sparse UNet takes the voxel grid along with its 3-dim features as input and outputs neural volumetric features, consisting of n-dim feature for each occupied voxel. This yields our Lidar embeddings $P = \{(\mathbf{p}_i, f_i)|i=1,...,N\}$, where each point $i$ is located at $\mathbf{p}_i$ and associated with a vector $f_i$, which is the neural feature of the voxel cell it resides in, encoding the local and global geometry around $\mathbf{p}_i$. 

\vspace{0.1cm}
\noindent \textbf{Query of Lidar features.} For each sample point $\textbf{x}$ along the ray to be rendered, we query its Lidar feature if there are at least $K$ nearby Lidar points within a search radius $R$; otherwise, its Lidar feature is set as empty (\ie all-zero). Specifically, we employ a Fixed Radius Nearest Neighbors (FRNN) approach~\cite{hoetzlein2014fast} to search a $K$-nearest Lidar point index set with respect to $\textbf{x}$, denoted as $\mathcal{S}_\textbf{x}^{K}$. 
Different from ~\cite{chang2023neural} where ray sampling is predetermined prior to initiating the training process, our method conducts FRNN searching online as the distribution of the sampled points from our proposal networks shift dynamically towards concentrating on the surface as the NeRF training converges. 

Following Point-NeRF~\cite{xu2022point}, our method harnesses an MLP, $\mathcal{F}$, to map Lidar feature from each point to a neural scene description. For the $i$-th point neighbor of $\mathbf{x}$, $\mathcal{F}$ takes as input the Lidar feature $f_{i}$ and relative position $\mathbf{x}{-}\mathbf{p}_i$, and outputs the neural scene description as 
\begin{equation}
f_{i, \mathbf{x}}=\mathcal{F}\left([f_i, \mathbf{x}-\mathbf{p}_i]\right),
\label{eq:offset_mlp}
\end{equation}
where $[,]$ indicates concatenation. To obtain the final Lidar encoding $\phi_L(\mathbf{x})$ at the sampled location $\mathbf{x}$, we use standard inverse-distance weighting to aggregate the neural scene description $f_{i, \mathbf{x}}$ from its $K$ neighboring points, 

\vspace{-1em}
\begin{align}
\phi_L(\mathbf{x}) &= \begin{cases}\frac{\sum_{i \in \mathcal{S}^{K}_{\textbf{x}}} w_{i} f_{i, \mathbf{x}}}{\sum_{i \in \mathcal{S}^{K}_{\textbf{x}}} w_{i}}, & \text{if}~ \mathcal{S}_\textbf{x}^{K} ~\text{is not}~\varnothing, \\ \mathbf{0}, & \text{otherwise.} \end{cases} \\
\text{and}~~~w_i &= \frac{1}{\left\|\mathbf{p}_i-\mathbf{x}\right\|}.
\end{align}
\label{eq:aggregate}
\vspace{-1em}

\vspace{0.1cm}
\noindent \textbf{Feature fusion for radiance decoding.} Different from~\cite{xu2022point,chang2023neural} that solely relies on point features, we concatenate the Lidar encoding $\phi_L$ with hash encoding $\phi_h$~\cite{mueller2022instant}, and apply an MLP $\mathcal{F}_{\alpha}$ to predict the per-sample density $\alpha$ and density embedding $\mathbf{h}$. Finally, the corresponding color $\mathbf{c}$ is predicted from the spherical harmonics encoding $\textit{SH}$ of the viewing direction $\mathbf{d}$ and density embedding $\mathbf{h}$, via another MLP $\mathcal{F}_{\mathbf{c}}$:

\vspace{-1em}
\begin{align}
\alpha, \mathbf{h} &= \mathcal{F}_{\alpha}([\phi_L(\mathbf{x}), \phi_h(\mathbf{x}])),\\
\mathbf{c} &= \mathcal{F}_{\mathbf{c}}([\mathbf{h}, \textit{SH}(\mathbf{d})]).
\end{align}
\label{eq:final}
\vspace{-1em}


%% file: sec/3_3_ds.tex
\subsection{Robust Depth Supervision}
\label{subsec:ds}
\input{figure/ds}

\noindent \textbf{Motivation.} In addition to the feature encoding, we also derive depth supervision from the Lidar points by projecting them onto the image plane. However, the sparse nature of Lidar points yields limited benefits, that are not sufficient for reconstructing low-texture regions such as road surface. Here, we put forth to accumulate adjacent Lidar frames to increase density. Despite the 3D points accurately capturing the scene structure, one needs to account for inter-points occlusion when projecting them onto the image plane for depth supervision. The occlusion arises due to the increased displacement between the camera and the Lidar from its adjacent frames, yielding bogus depth supervision, as illustrated in \cref{fig:ds}. This is non-trivial to handle due to the sparsity of Lidar even after accumulation, making principled graphic techniques such as z-buffering not applicable. In this work, we propose a robust supervision scheme to automatically filter out bogus depth supervision in while training NeRF. 

\vspace{0.1cm}
\noindent \textbf{Occlusion-aware robust supervision scheme.} 
We design a curriculum training strategy such that the model initially trains with closer, more reliable depth data, which are less prone to occlusion. As training progresses, the model gradually begins to incorporate more distant depth data. Concurrently, the model develops the capacity to discard depth supervisions that are anomalously distant compared to its predictions. Formally, with the pool of all depth points denoted as $\mathcal{D}=\{\mathcal{D}_1,\mathcal{D}_2,...,\mathcal{D}_N\}$, and further denoting the NeRF-rendered depth corresponding to $\mathcal{D}_i$  as $\hat{\mathcal{D}}_i$, we identify the reliable subset of depth points $\mathcal{D}^{m}_{\text{reliable}}$ in the $m$-th training iteration as:
\begin{align}
    \mathcal{D}^{m}_{\text{reliable}} &= \{\mathcal{D}_{i}~|~\mathcal{D}_{i} {\leq} \epsilon^{m}_t,~~\mathcal{D}_{i}  \leq \hat{\mathcal{D}}_{i} {+} \epsilon^{m}_o, ~~\mathcal{D}_{i} {\in} \mathcal{D}\}, \\
    \epsilon^{m}_{t} &= \text{min}\{\alpha_t \epsilon^{m-1}_{t}, ~\epsilon_{t} \},~~~~~\alpha_t>1, \\
    ~\epsilon^{m}_{o} &= \text{max}\{\alpha_o \epsilon^{m-1}_{o}, ~\epsilon_{o} \},~~~~\alpha_o<1.
\end{align}
\label{eq:robust_set}
One notices that $\mathcal{D}^{m}_{\text{reliable}}$ is governed by two scheduled parameters: valid depth threshold $\epsilon^{m}_t$ and valid depth offset $\epsilon^{m}_o$. The $\epsilon^{m}_t$ serves to filter out depth samples exceeding this threshold,  thereby prioritizing nearer depth samples which are less likely to be occluded. 
As training progresses, $\epsilon^{m}_t$ is exponentially increased at a rate of $\alpha_t$ to involve more depth supervision from further field. Meanwhile, samples exhibiting a depth value far larger than the predicted depth $\hat{\mathcal{D}}_{i}$ are omitted, as they are likely occluded points. This is thresholded by $\hat{\mathcal{D}}_{i}+\epsilon^{m}_o$, with  $\epsilon^{m}_o$ decaying exponentially at a rate of $\alpha_o$, in tandem with the improvement of depth predictions over the course of training.  

\input{figure/cdf_loss}

\paragraph{Lidar Depth Loss} For samples in $\mathcal{D}^{m}_{\text{reliable}}$, we adopt the pixel-level depth loss proposed in URF~\cite{rematas2022urban}, written as $\mathcal{L}_{\text{ds}} = \mathcal{L}_{\text{depth}} + \mathcal{L}_{\text{sight}}$. In addition to a $L_2$ loss $\mathcal{L}_{\text{depth}}$ between the rendered depth and the ground truth, a line-of-sight prior $\mathcal{L}_{\text{sight}}$ 
is applied to further constrain each sampling point individually. But in contrast an approximate computation of $\mathcal{L}_{\text{sight}}$ in NeRFstudio, we implement an exact one. Specifically, we first note that the volume rendering boils down to a weighted sum of the predicted color on sampled points along the ray (see supplementary for equations). Given the weight should ideally concentrate around surfaces, $\mathcal{L}_{\text{sight}}$ enforces the weight distribution to resemble a Gaussian distribution $\mathcal{N}(\hat{\mathcal{D}}_i,\epsilon_n)$ centered at the ground truth depth $\hat{\mathcal{D}_i}$ along the rendering ray, written as 
\begin{equation}
    \mathcal{L}_{\text{sight}} = \mathbb{E}_{\mathcal{D}_i \in \mathcal{D}^{m}_{\text{reliable}}}\left[ \int_{t_{near}}^{t_{far}} (w(t)-\mathcal{N}(\hat{\mathcal{D}_i},\epsilon_n))^2 \,dt \right],
\end{equation}
where $w$ indicates the weight to be integrated along the distance $t$ on the rendering ray from $t_{near}$ to $t_{far}$. Since the weight $w$ is computed on discrete intervals given by point sampling in NeRF, this loss is discretized to

\begin{equation}
    \mathcal{L}_{\text{sight}} = \mathbb{E}_{\mathcal{D}_i \in \mathcal{D}^{m}_{\text{reliable}}}\left[\sum_i (w_i-\mathcal{N}_i)^2 \right], 
\end{equation}
where $\mathcal{N}_i$ indicates the probability mass within the $i$-th interval. Here, a possible implementation (\eg in Nerfstudio~\cite{nerfstudio}) to obtain $\mathcal{N}_i$ is by mid-point approximation as illustrated in \cref{fig:cdfloss}. However, we note that this approximation is  unnecessary and implement differently based on cumulative distribution function (CDF) -- the probability mass of a Gaussian distribution can be obtained through its tabulated CDF. We show in supplementary that our exact implementation leads to improved PSNRs.

%% file: figure/ds.tex
\begin{figure}
  \centering
  \includegraphics[width=1.0\linewidth]{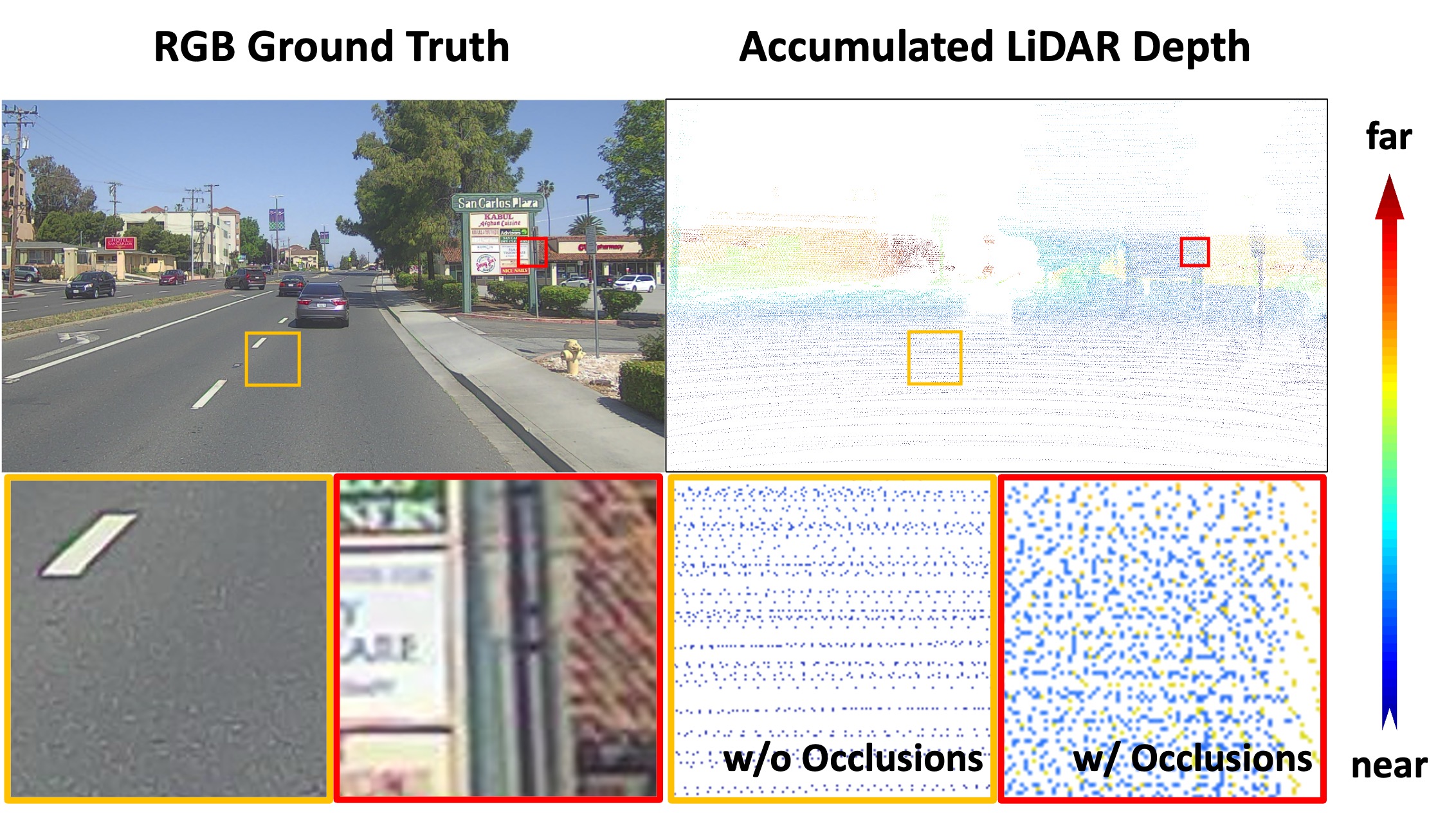}
  \centering
  \caption{
  \textbf{Illustration of the occlusion issue on the depth map projected from accumulated Lidar points.} Observe that multiple layers of surface points may project to the same region on the image, yielding ghost depth points.
  }
  \label{fig:ds} 
  \vspace{-1em}
\end{figure}

%% file: figure/cdf_loss.tex
 \begin{figure}[t!]
  \centering
  \includegraphics[width=1.0\linewidth]{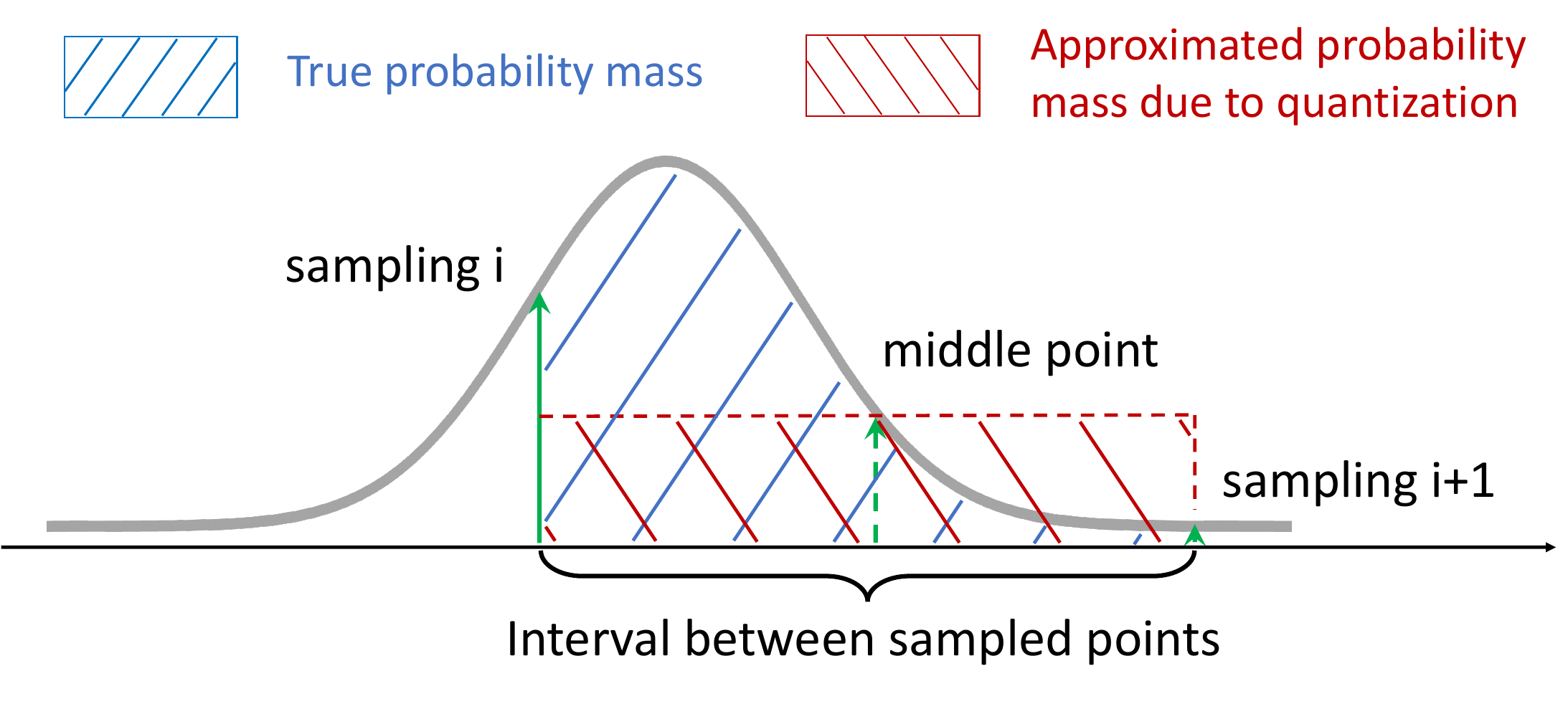}
  \centering
  \vspace{-2em}
  \caption{Illustration of the true probability mass and its mid-point approximation.}
  \label{fig:cdfloss} 
  \vspace{-1em}
\end{figure}

%% file: sec/3_5_aug.tex
\subsection{Augmented View Supervision}
\label{subsec:avs}

Recall that a vehicle-mounted camera generates sparse training images with limited view coverage due to its forward motion, posing challenges for NeRF reconstruction, especially when the novel views deviate from vehicle trajectory. Here, we propose to augment training data leveraging Lidar. First, we colorize the point clouds in each Lidar frame  by projecting onto its synchronized camera and interpolate the image for RGB values. The colorized point clouds are accumulated as described in \cref{subsec:ds} and projected onto a set of synthetically augmented views, yielding synthesized images and depth maps as illustrated in \cref{fig:pipeline}.  These augmented training views are derived from existing ones, by introducing stochastic perturbations to their camera centers, with the shifting magnitude $\epsilon_a{\in}\mathcal{N}(0, \epsilon_a)$. Nonetheless, such augmented data fails to account for potential occlusions as depicted in \cref{subsec:ds}.  Our model, fortified by robust depth supervision, is adept at discerning and excluding occluded Lidar points online. The augmented views are used to train NeRF similarly as the real training views, and we denote the extra loss separately as $\mathcal{L}_{aug}$.


%% file: sec/4_experiments.tex
\section{Experiments}
\label{sec:exp}
\input{tab/comp}

In this section, we detail our experimental setup and benchmark our method against state-of-the-art NeRF techniques, demonstrating superior photorealism. We further ablate our design choices, underscoring the effectiveness of robust depth supervision, LiDAR encoding, and augmented view supervision in enhancing realism.

\input{sec/4_1_setup}

\input{sec/4_2_nvs}

\input{sec/4_3_ablation}

\subsection{Results on NuScenes and Argoverse}

\input{tab/comp_nuscenes}
\input{figure/vis_comp_nuscenes}

we evaluate \Ours on the NuScenes dataset with comparison to
S-NeRF, by adapting their method to using the single front camera. We present quantitative results in Tab.~\ref{tab:comp_nuscenes} and qualitative example in Fig.~\ref{fig:vis_comp_nuscenes}, both showing superior performance from our method over S-NeRF. Ablation study results are also shown in Tab.~\ref{tab:comp_nuscenes}, which further consolidate the efficacy of our proposed components.

\input{figure/vis_comp_argoverse}

Next, we evaluate on Argoverse dataset following the protocol in \cite{chang2023neural}. We present example qualitative comparisons in \cref{fig:argoverse} while leaving more evaluations to the supplementary. As can be seen, our \Ours achieves far superior rendering quality with high resolution. We note the \cite{chang2023neural}'s blank rendering on regions not covered by Lidar, as they solely rely on Lidar for radiance decoding. This illustrates the advantage of our framework in combining the complementary benefits from Lidar encoding and high-resolution hash grid. 


%% file: tab/comp.tex
\begin{table}[]
    \centering
    \resizebox{1.0\linewidth}{!}{
    \begin{tabular}{lccccc}
    \toprule
    \multirow{2}{*}{Methods} & \multicolumn{3}{c}{Interpolation} &
    \multicolumn{2}{c}{Lane Shift} \\
    \cmidrule(r){2-4} \cmidrule(l){5-6}
    & {PSNR$\uparrow$ } & {SSIM$\uparrow$ } & {LPIPS$\downarrow$ } & {FID$\downarrow $ @ 2m} & {FID$\downarrow $ @ 3.7m} \\
    \midrule
    Instant-NGP & 24.282  & 0.733 & 0.408  & 140.3 & 173.2 \\
    Mip-NeRF 360 & 23.693 & 0.691 & 0.496 & 189.4 & 231.1 \\
    Nerfacto & 27.122 & 0.804 & 0.268 & 116.7 & 151.0 \\
    UniSim & 26.014 &  0.768 & 0.342 & 118.5 & 141.3 \\
    \Ours(Ours) & {27.255} & \textbf{0.812} & \textbf{0.224} & \textbf{106.5} & \textbf{126.0} \\
    \bottomrule
    \end{tabular}}
	\caption{
            \textbf{Quantitative comparisons on view synthesis} with state-of-the-art NeRF variants. 
        }
	\label{tab:comp}
\vspace{-1em}
\end{table}

%% file: sec/4_1_setup.tex
\subsection{Experimental Setup}
\label{subsec:setup}

\noindent \textbf{Datasets.}
Following~\cite{yang2023unisim}, we rely on \textbf{Pandaset}~\cite{xiao2021pandaset} as the primary dataset for evaluation, using its front camera and synchronized spinning Lidar. Each scene is consisting of 80 frames captured at 10Hz. 
We leverage its sensor localization for Lidar accumulation.
Since our focus is on static scenes,  
the dynamic vehicles are masked out during evaluation, similarly as in~\cite{chang2023neural}.
We also evaluate on \textbf{NuScenes}~\cite{caesar2020nuscenes} and \textbf{Argoverse2}~\cite{wilson2023argoverse} to compare with S-NeRF~\cite{xie2022s} and NeRF-LiDAR-cGAN~\cite{chang2023neural}, respectively. 

\vspace{0.1cm}
\noindent \textbf{Baselines.}
On Pandaset, we compare our model against several modern implicit-based neural radiance fields methods: \textbf{Instant-NGP}, \textbf{Mip-NeRF 360}, \textbf{Nerfacto} and \textbf{UniSim}.
Instant-NGP~\cite{mueller2022instant} adopts multi-resolution hashing encoding for compact scene representation and efficient rendering. Mip-NeRF 360 \cite{barron2022mip} adopts integrated position encoding with scene contraction for handling unbounded scenes.
Nerfacto~\cite{nerfstudio} combines the compact representation from Instant-NGP and proposal network from Mip-NeRF 360. 
UniSim~\cite{yang2023unisim} is the state-of-the-art simulator for street scenes, reconstructing both the static background and dynamic actors with neural feature grids. In our experiments, we mainly focus on modeling the static background.

\vspace{0.1cm}
\noindent \textbf{Implementation Details.}
We mask dynamic objects in RGB images using dataset bounding box annotations and an instance segmentation model~\cite{he2017mask}. Static Lidar points are isolated by omitting points within dynamic objects' 3D bounding boxes. For nearest neighbor searches, we use a CUDA-based FRNN search algorithm~\cite{githubGitHubLxxueFRNN} to query $K{=}6$ nearest Lidar points within a $0.3m$ radius. Our loss weights and scheduling parameters in depth training scheme are given in supplementary.
Instant-NGP, Nerfacto, UniSim and our proposed \Ours use identical size of hash grid and hidden layers in MLPs. 
Additional details are in the supplementary.

%% file: sec/4_2_nvs.tex
\subsection{Novel View Synthesis Results}
\label{subsec:nvs}
\input{figure/vis_comp}

In our experiments, we assess novel view synthesis under interpolation and extrapolation settings. For interpolation setting, we randomly subsample RGB and Lidar frames, testing on every fourth frame and training on the rest. We follow common practice to report PSNR, SSIM, and LPIPS interpolation views. For extrapolation setting, following~\cite{yang2023unisim}, we simulate new trajectories by laterally shifting them left or right by 2 or 3.7 meters (lane width of Interstate Highway standards~\cite{wikipediaInterstateHighway}). We report FID at the perceptual level since ground-truth is unavailable,

As shown in \cref{tab:comp}, our method outperforms all others in every metric. While methods such as Nerfacto and UniSim show robust performance in the interpolation setting, Mip-NeRF 360 lags behind. The qualitative gap becomes even more significant in extrapolation scenarios. \cref{fig:vis_comp} displays these qualitative differences, where our method exhibits enhanced visual realism compared to the baselines, particularly in rendering fine structures, thanks to our LiDAR encoding. This is especially notable as, even though UniSim also utilizes LiDAR depth supervision, our method more effectively renders low-texture areas like road surfaces, demonstrating the advantage of our deeper utilization of Lidar.

%% file: figure/vis_comp.tex
\begin{figure*}
\vspace{-1em}
  \centering
  \includegraphics[width=0.95\linewidth]{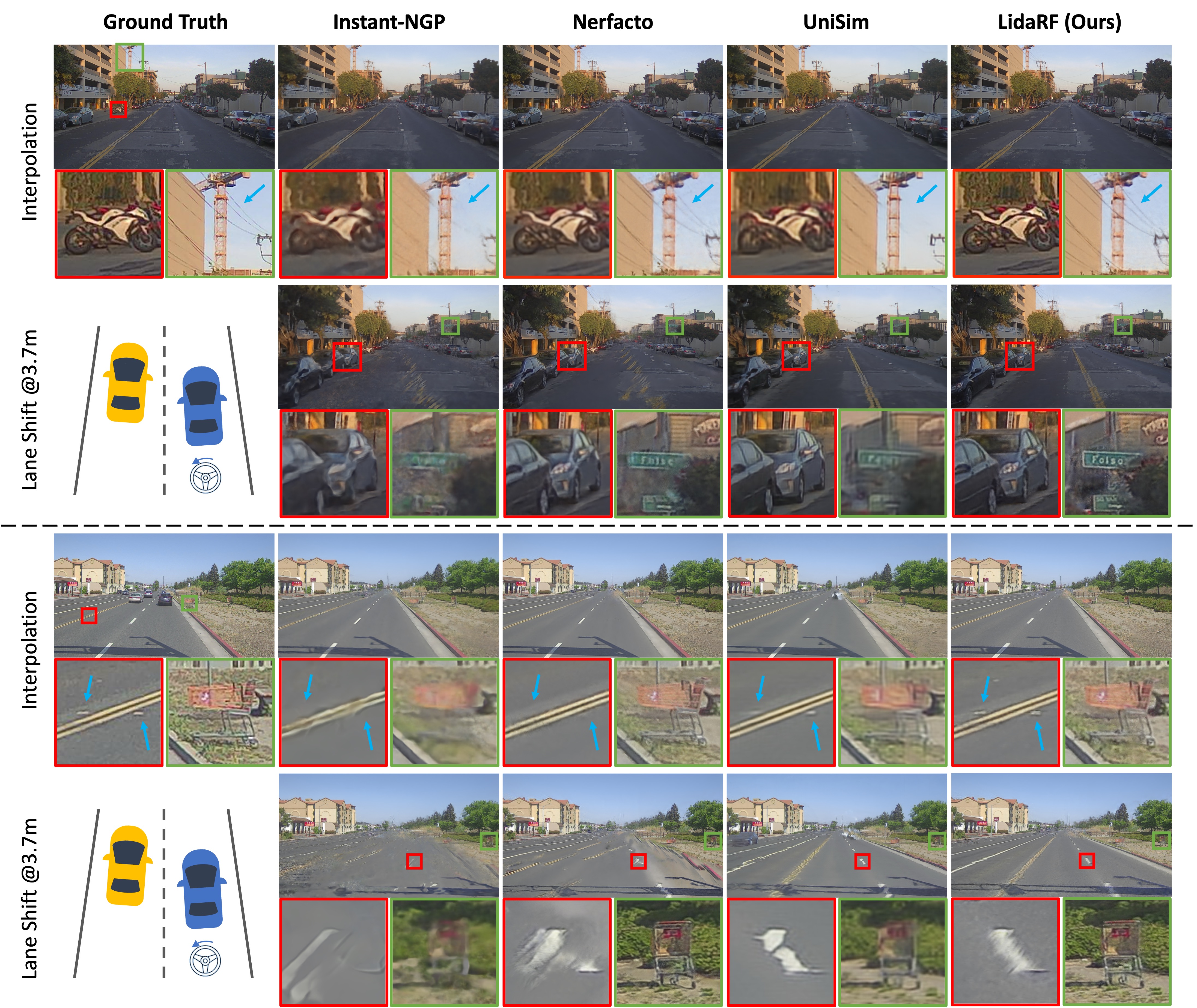}
  \centering
  \vspace{-0.5em}
  \caption{
  \textbf{Qualitative Comparison} on novel view synthesis from different methods. We evaluate on both the interpolation and extrapolation views, the latter of which corresponds to a lane shift. We highlight the performance gap with boxes.  
  }
  \label{fig:vis_comp} 
\vspace{-1em}
\end{figure*}

%% file: sec/4_3_ablation.tex
\subsection{Ablation Study}
\label{subsec:ablation}
\input{tab/ablation}

In our ablation study presented in Tab.~\ref{tab:ablation}, we evaluate the impact of three key components -- robust depth supervision, Lidar encoding, and augmented view supervision. The baseline for comparison, shown in the first row of Table~\ref{tab:ablation}, is the Nerfacto method. We denote different depth supervision strategies as follows: $\mathcal{L}^{1}_{ds}$ for single-frame Lidar depth supervision, $\mathcal{L}^{10}_{ds}$ for depth map supervision using 10 adjacent Lidar frames, and $\mathcal{L}_{ds}$ for the same but within our robust supervision scheme.  
As another baseline for occlusion handling, we apply hidden point removal (HPR) algorithm~\cite{katz2007direct} implemented in Open3D~\cite{Zhou2018} to remove occluded points as data preprocessing; the supervision after HPR is denoted as $\mathcal{L}^{\text{HPR}}_{\text{ds}}$. Lidar encoding is represented by $\phi_L$.
Lastly, $\mathcal{L}_{\text{aug}}$ signifies supervision with augmented RGB and depth data derived from Lidar projections.

\input{figure/ds_ablation}
\vspace{0.1cm}
\noindent \textbf{Effects of Robust Depth Supervision.}
In Tab.~\ref{tab:ablation}, we illustrate that Lidar depth markedly improves LPIPS and FID in lane shift settings with marginal PSNR drop compared to Nerfacto. Notably, our robust supervision scheme with accumulated Lidar points further enhances all metrics. Also note the offline HPR significantly lags behind our more adaptive, NeRF-informed scheme for occlusion handling. Fig.~\ref{fig:ds_ablation} compares different Lidar depth supervision settings. 
Under the supervision of $\mathcal{L}^{1}_{\text{ds}}$, which utilizes single-frame Lidar depth maps known for their sparsity and reduced occlusions, the rendered depths exhibit high accuracy in texture-rich areas (as indicated by the yellow boxes). However, this accuracy significantly diminishes in regions with thin structures, due to a lack of abundant geometric guidance, as observed in the red boxes. 
Conversely, the supervision with $\mathcal{L}^{10}_{\text{ds}}$, involving with noisy Lidar depth maps accumulated from 10 adjacent frames, leads to rendered depths that display noticeable noise. This is attributed to the prevalent depth ambiguities within the data. 
Employing our proposed $\mathcal{L}^{\text{robust}}_{ds}$, our methodology effectively leverages denser depth maps for the precise reconstruction of intricate structures (highlighted in red boxes), as well as excludes some occluded depths (noted in yellow boxes).
See more examples in the supplementary materials.

\vspace{0.1cm}
\input{figure/lenc_ablation}
\noindent \textbf{Effects of Lidar Encoding.}
From Tab.~\ref{tab:ablation}, it is evident that Lidar encoding contributes to enhancements across all metrics. As shown in Fig.~\ref{fig:lenc_ablation}, Lidar encoding enables our method to produce sharper textures.
This enhancement stems from our sparse convolution-based architecture, which is resilient to Lidar point noise and density variations. 

\vspace{0.1cm}
\input{figure/aug_view_ablation}
\noindent \textbf{Effects of Augmented View Supervision.}
While the quantitative benefits of augmented view supervision, as per Tab.~\ref{tab:ablation}, appear modest, it consistently enhances performance in extrapolation scenarios across all test scenes. This is particularly notable in scenes where other methodologies fall short. Fig.~\ref{fig:aug_ablation} showcases an instance of this: in scenes with parking cars that are scarcely captured in the raw training data due to the rapid movement of the camera vehicle, our augmented view supervision significantly elevates the quality.

%% file: tab/ablation.tex
\begin{table}
\begin{center}
\resizebox{1.0\linewidth}{!}{
\begin{tabular}{ p{0.2cm} p{0.2cm}  p{0.2cm} p{0.2cm} p{0.2cm} p{0.2cm}  cccc } 
 \rotatebox{90}{$\mathcal{L}^1_{\text{ds}}$} & \rotatebox{90}{$\mathcal{L}^{10}_{\text{ds}}$} & \rotatebox{90}{$\mathcal{L}^{\text{HPR}}_{\text{ds}}$} & \rotatebox{90}{$\mathcal{L}^{\text{robust}}_{\text{ds}}$} &  \rotatebox{90}{$\phi_L$} &  \rotatebox{90}{$\mathcal{L}_{\text{aug}}$} & PSNR$\uparrow$  &  SSIM$\uparrow$ & LPIPS$\downarrow$   & FID$\downarrow$ @3.7m\\
\hline
\untickbox & \untickbox & \untickbox & \untickbox & \untickbox & \untickbox & 27.122 & 0.804 & 0.268 & 151.0  \\
\tickbox & \untickbox & \untickbox & \untickbox & \untickbox & \untickbox   & 27.016 & 0.800 & 0.264 & 138.2 \\
\untickbox & \tickbox & \untickbox & \untickbox & \untickbox & \untickbox   & 26.946 & 0.797	 & 0.264 & 137.7 \\
\untickbox & \untickbox & \tickbox & \untickbox & \untickbox & \untickbox   & 27.000 & 0.799	 & 0.261 & 139.1 \\
\hdashline

\untickbox & \untickbox & \untickbox & \tickbox &  \untickbox & \untickbox & 27.090 & 0.804  & 0.247 & 131.7 \\
\untickbox & \untickbox & \untickbox & \tickbox & \tickbox & \untickbox  & 27.219 & 0.810  & 0.228 & 128.7 \\
\untickbox & \untickbox & \untickbox & \tickbox & \tickbox & \tickbox   & \textbf{27.254} & \textbf{0.812}  & \textbf{0.223}  & \textbf{126.0} \\ \bottomrule
\end{tabular}}
\end{center}
\vspace{-1em}
\caption{
\textbf{Ablation study} of our robust depth supervision, Lidar encoding, and training view augmentation on Pandaset.
}
\label{tab:ablation}
\vspace{-1em}
\end{table}

%% file: figure/ds_ablation.tex
\begin{figure*}
  \centering
  \includegraphics[width=1.0\linewidth]{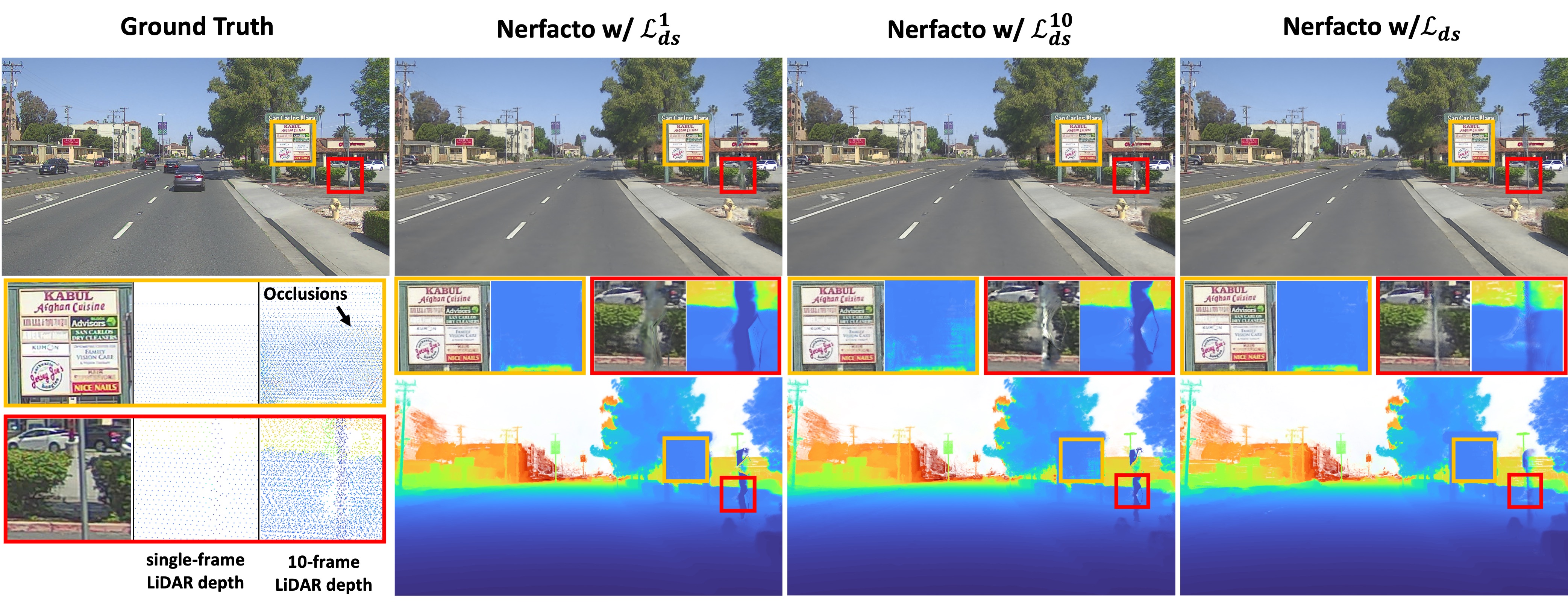}
  \centering
  \vspace{-2em}
  \caption{
  \textbf{Qualitative comparison between different LiDAR Depth Supervisions} -- With sparse depths, Nerfacto w/ $\mathcal{L}^{i}_{\text{ds}}$ fails to model thin structures. With denser but noisy depths, Nerfacto w/ $\mathcal{L}^{i}_{\text{ds}}$ generate ambiguous depth predictions. Our proposed schemes is robust to occlusions and able to learn delicate structures.
  Our proposed depth supervision scheme can learn delicate structures with noisy depth map, while the other settings both fail.
  }
  \label{fig:ds_ablation} 
 \vspace{-1em}
\end{figure*}

%% file: figure/lenc_ablation.tex
\begin{figure}
  \centering
  \includegraphics[width=1.0\linewidth]{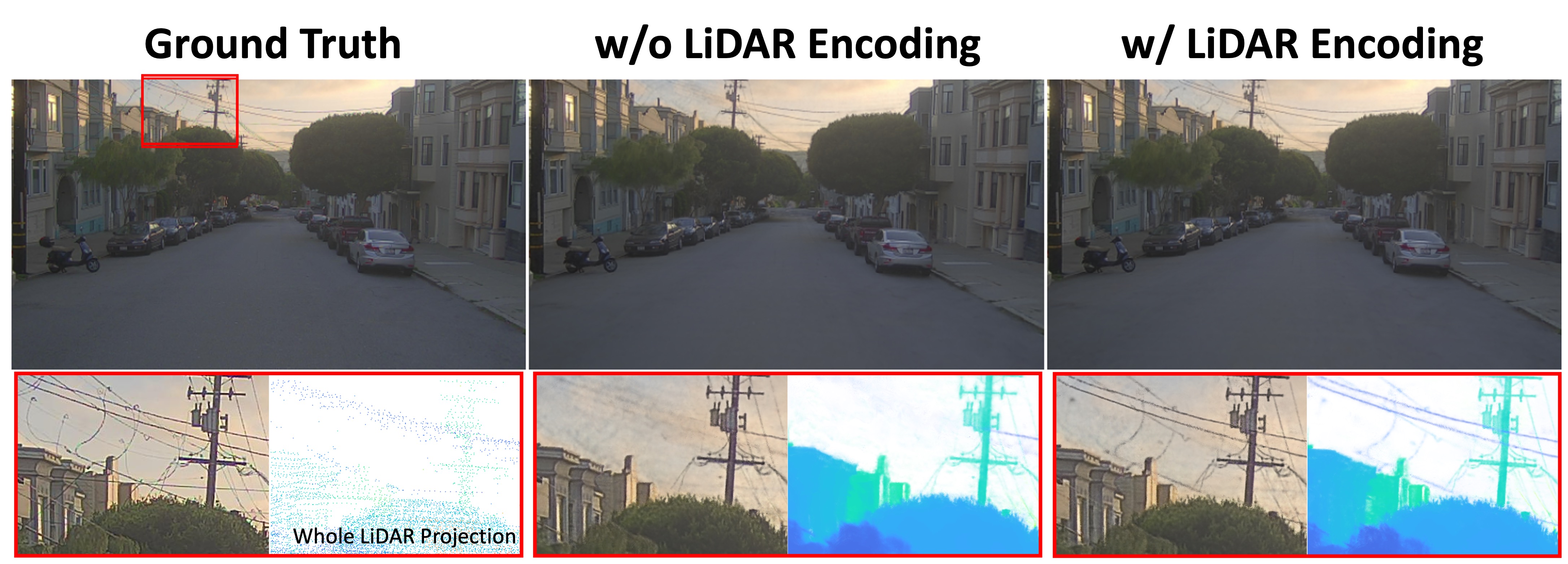}
  \centering
  \vspace{-2em}
  \caption{
  \textbf{Qualitative comparison w.r.t Lidar encoding} -- Lidar encoding is beneficial in modelling sharp textures, \eg power lines.
  }
  \label{fig:lenc_ablation} 
\vspace{-1em}
\end{figure}


%% file: figure/aug_view_ablation.tex
\begin{figure}
  \centering
  \includegraphics[width=1.0\linewidth]{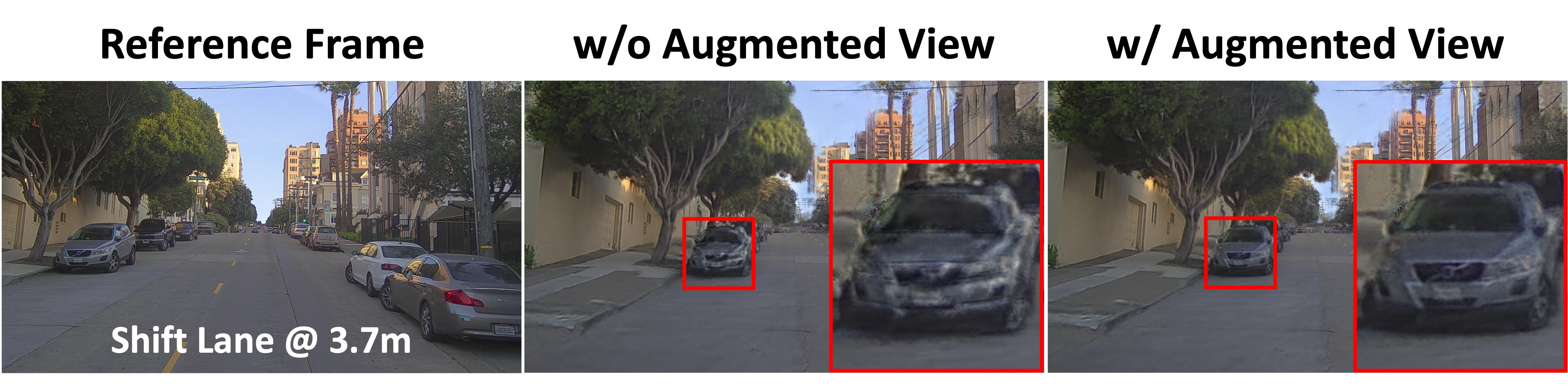}
  \centering
  \vspace{-2em}
  \caption{
  \textbf{Qualitative comparison w.r.t augmented view supervision} -- it largely improves rendering quality on the regions which are scarcely captured in the raw training data.
  }
  \label{fig:aug_ablation} 
\vspace{-0.5em}
\end{figure}

%% file: tab/comp_nuscenes.tex
\begin{table}[t]
    \centering
    \resizebox{0.95\linewidth}{!}{
    \begin{tabular}{cccccc}
    \toprule
    \multirow{2}{*}{Methods} & \multirow{2}{*}{S-NeRF} &
    \multicolumn{4}{c}{\Ours(Ours)} \\
    \cmidrule{3-6}
    &  & w/o $\mathcal{L}_{\text{ds}}$ & w/o $\phi_L$ & w/o $\mathcal{L}_{\text{aug}}$ & Full  \\
    \midrule    
    {PSNR$\uparrow$ } & 29.377  & 30.629  & 31.001 & 31.133 & \textbf{31.162} \\
    {SSIM$\uparrow$ } & 0.859 & 0.871 & 0.873 & 0.883 & \textbf{0.884} \\
    {LPIPS$\downarrow$ } & 0.349 & 0.278 & 0.237 & 0.222 & \textbf{0.211} \\
    \bottomrule
    \end{tabular}}
    \vspace{-0.5em}
    \caption{Evaluation on NuScenes with comparison to S-NeRF.}
\label{tab:comp_nuscenes}
\end{table}

%% file: figure/vis_comp_nuscenes.tex
\begin{figure}[t]
  \centering
  \includegraphics[width=1.0\linewidth]{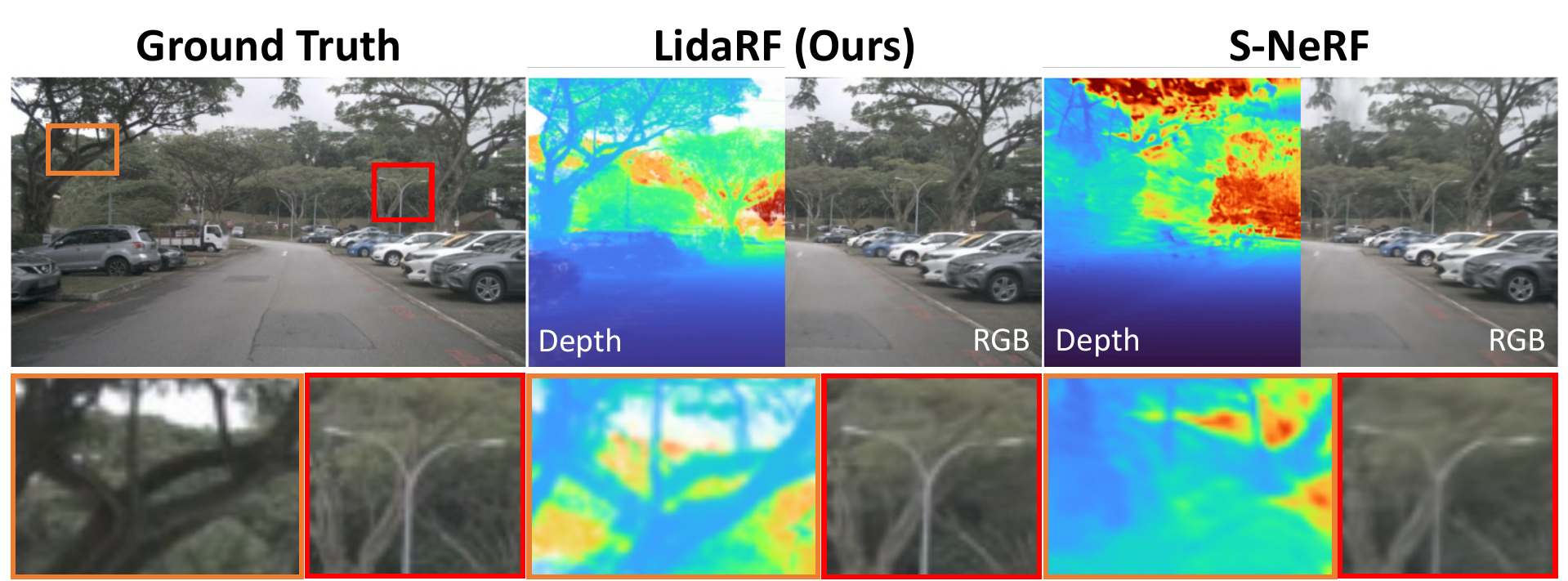}
  \centering
  \vspace{-2em}
  \caption{Visual Comparison with S-NeRF on NuScenes.}
  \vspace{-1em}
  \label{fig:vis_comp_nuscenes}   
\end{figure}

%% file: figure/vis_comp_argoverse.tex
\begin{figure}[t!]
  \centering
  \includegraphics[width=1.0\linewidth]{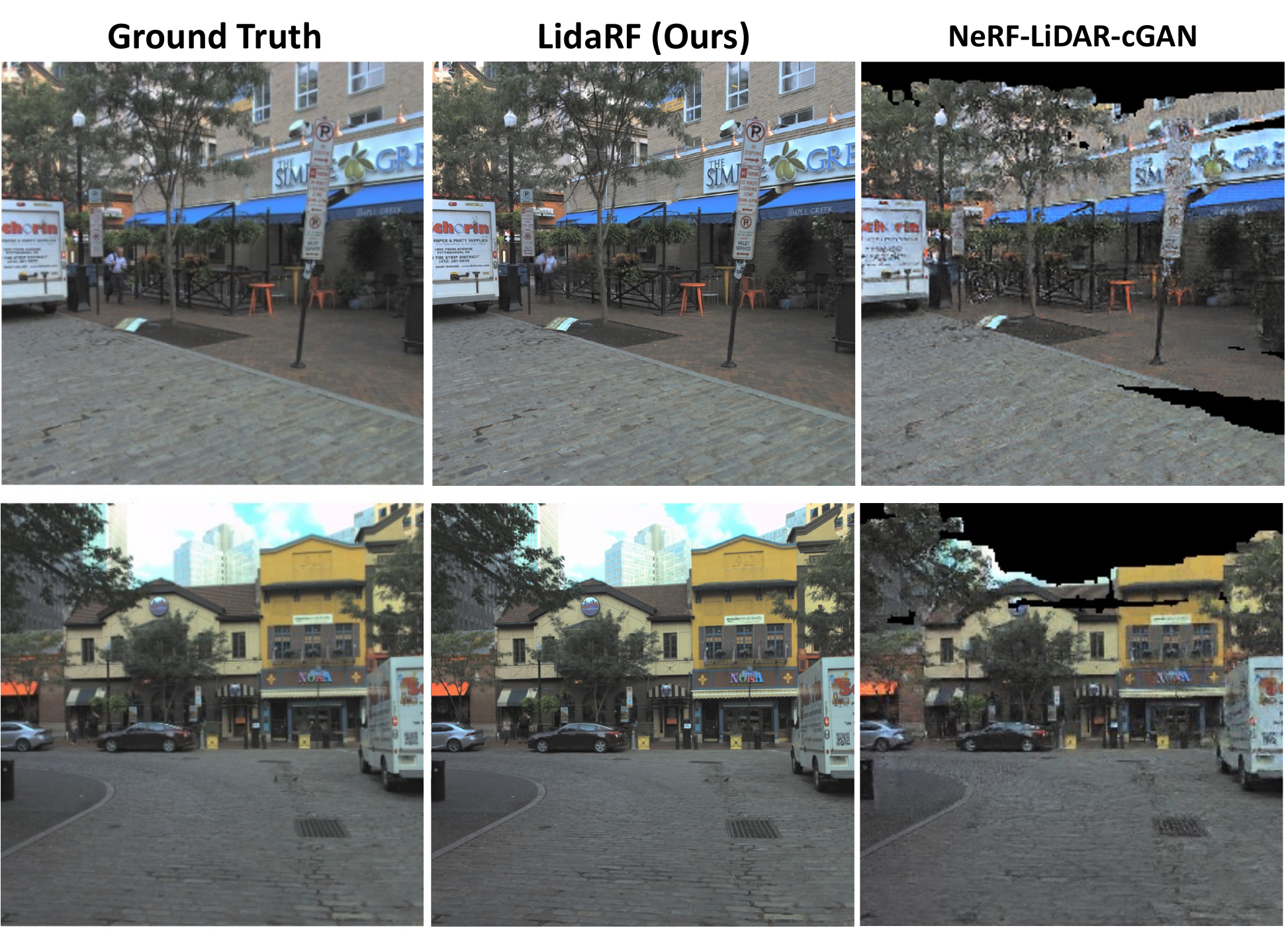}
  \centering
  \vspace{-1.5em}
  \caption{\textbf{Qualitative results on Argoverse} with comparison to NeRF-LiDAR-cGAN~\cite{chang2023neural} that solely relies on Lidar encoding, unlike our hybrid scene representation.}
  \label{fig:argoverse} 
\vspace{-1em}
\end{figure}

%% file: sec/6_conclusion.tex
\section{Conclusion}
\label{sec:conclusion}
In this paper, we focus on unlocking the potential of Lidar to improve NeRF on road scenes, which remain a challenging scenario for novel view synthesis due to highly constrained camera motions. We develop insights on fusing Lidar encoding with high-resolution grid based representation to reap their complementary benefits, and further, extract more robust and extensive depth supervision from Lidar. A limitation of our work is that we currently handle static background only. We envision that the key insights developed in this paper can benefit dynamic objects as well, which remains an interesting future direction to explore. 


%% file: sec/X_suppl.tex
\section{Additional Ablation Results}
\label{sec:supp_ablation}

\noindent \textbf{Lidar Encoding.} Here, we discuss additional experiments to study the advantage of our Lidar encoding based on the 3D sparse convolutional network. Firstly, in view of the increased representation capacity brought by Lidar encoding,
we evaluate the impact of naively increasing the hash grid feature size without using any Lidar encoding, specifically doubling it from two to four features per level, denoted as ``Double Hash" in Table~\ref{tab:lidar_encoding_ablation}. This modification yields improved results in interpolation settings but exhibits a marginal performance decline in lane shift scenarios, indicative of potential overfitting. 

Furthermore, instead of 3D convolutional network, we explore LiDAR encoding utilizing MLPs (as applied in \cite{chang2023neural}) and PointNet++\cite{qi2017pointnet++}. For the former, the MLP contains three hidden layers with feature size (64, 96, 128), and a final layer outputting a 64-dim feature vector. For PointNet++, the hidden feature dimensions mirror those used in our sparse UNet-based method, and its point encoder samples 4096, 1024, 256, and 56 points at different levels. To enhance the efficiency of its farthest point sampling and neighbor point grouping, we utilize CUDA-based implementations from PyTorch3D~\cite{ravi2020accelerating}. The results, as outlined in Table~\ref{tab:lidar_encoding_ablation}, indicate that Lidar encoding with MLP and PointNet++ underperforms our encoding with sparse UNet. This indicates the benefits brought about by learning from a more global context with the 3D convolutional network, which has proven a powerful backbone widely applied in state-of-the-art 3D perception frameworks~\cite{liu2023bevfusion,yin2021center}.

Lastly, the memory and time overhead incurred by our LiDAR encoding module is modest. Specifically, compared to Nerfacto with robust depth supervision, our LiDAR encoding introduces an additional memory usage of 1846MB and an incremental time cost of 0.1s per training iteration. Notably, during inference, this overhead is further reduced as LiDAR encoding is required only once per scene, rather than per batch.


\vspace{0.2cm}
\noindent \textbf{Robust Depth Supervision.}
\input{figure/ds_ablation_supp}
In Fig.~\ref{fig:ds_ablation_supp}, we present visual comparisons of different depth supervision settings under the lane shift scenario. With sparse or noisy depth supervision, Nerfacto w/ $\mathcal{L}^{1}_{\text{ds}}$ and Nerfacto w/ $\mathcal{L}^{10}_{\text{ds}}$ fail to model thin structures such as light poles.
In contrast, our proposed scheme is robust to occlusions and able to learn delicate structures with noisy depth maps.

\vspace{0.2cm}
\noindent \textbf{Augmented View Supervision.}
\input{figure/aug_view_ablation_supp}
Here, we provide more analyses on the impact of augmented view supervision. As shown in Fig.~\ref{fig:aug_ablation_supp}(a), we observe good performance with around 80-320 synthetic views, beyond which performance drops as it may dominate real training views. We randomly perturb the original views with Gaussian noise, and observe good performance with the standard deviation $\sigma$ around 1.0-1.5 meters, as shown in Fig.~\ref{fig:aug_ablation_supp}(b). We did not observe noticeable benefits from adding new orientations, likely as vehicles are mostly in forward motion.

\vspace{0.2cm}
\noindent \textbf{CDF and Mid-point Approximation.}
As mentioned in the main paper, Tab.~\ref{tab:cdf_loss_ablation}  provides a quantitative comparison of two implementations of LiDAR depth loss using single-frame LiDAR points. We observe improved performance from our exact implementation based on CDF, in comparison to the mid-point approximation. Despite  the marginal gap, they are consistently observed across all tested sequences in the Pandaset, particularly in the interpolation setting.

\section{Quantitative Results by Range}
\input{tab/psnr_by_range}
We perform more detailed evaluations by separating pixels into different depth ranges. In Tab.~\ref{tab:psnr_by_range}, we evaluate under two strategies for range separation -- 1) group the pixels with matching Lidar points into different distance ranges; 2) group pixels into foreground, sky, and their boundary region that is prone to artifacts due to discontinuity. The quantitative results indicate superior performance from our approach regardless of the range, compared to UniSim. That said, our proposed components are compatible to UniSim and may be combined in future work. 

\input{tab/lidar_encoding_ablation}
\input{tab/argoverse}

\section{Experiments on Argoverse}
\begin{figure*}[t!]
  \centering
  \includegraphics[width=0.65\linewidth, trim = 10mm 20mm 10mm 0mm, clip]{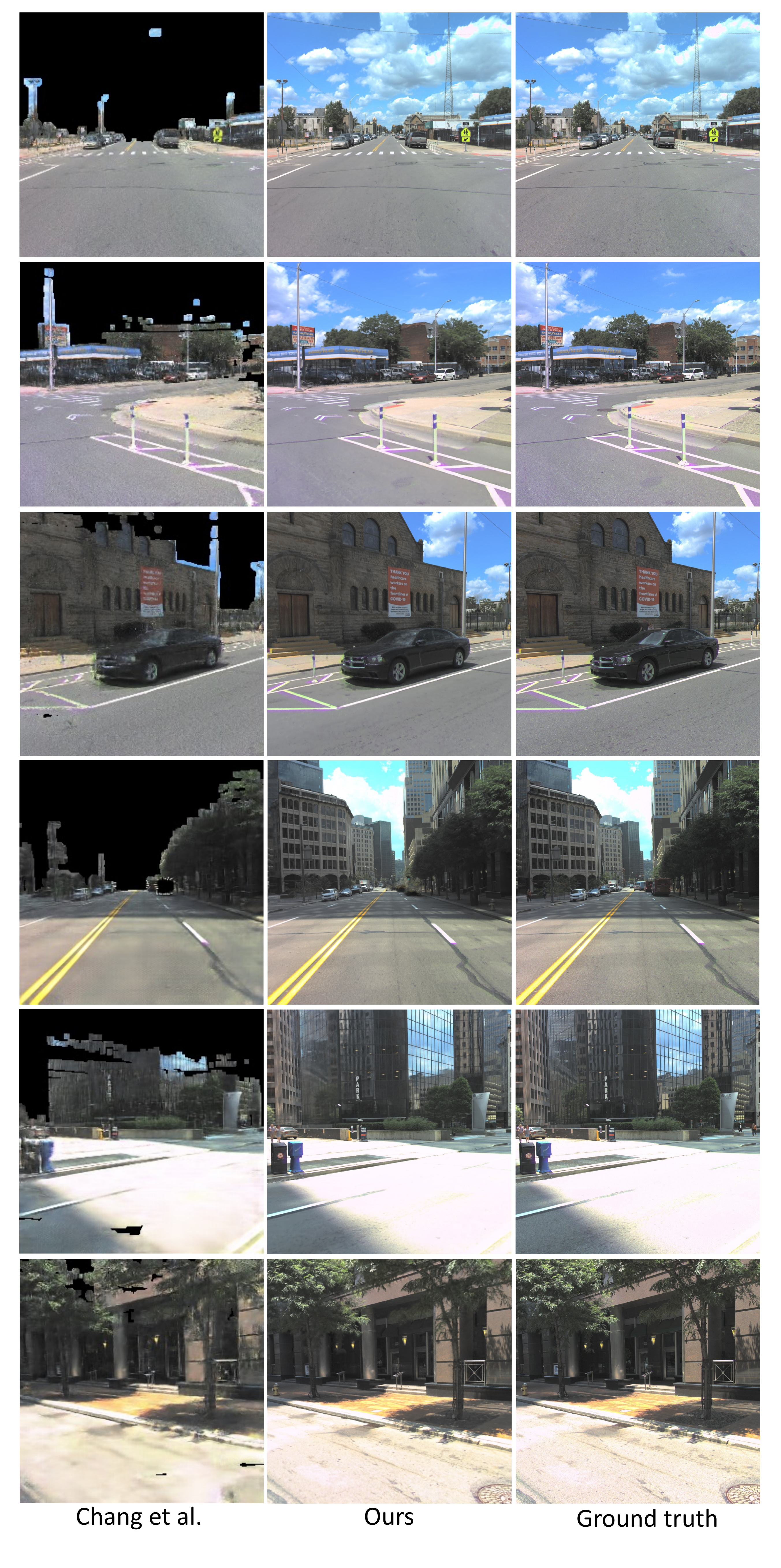}
  \centering
  \caption{\textbf{Qualitative comparison on the Argoverse dataset.} Our results are complete and of significantly higher resolution in comparison to those from Chang et al.~\cite{chang2023neural}.  }
  \label{fig:argoversequalitative_supp} 
  \vspace{-1em}
\end{figure*}

\input{tab/cdf_depth_loss}

As mentioned in the main paper, in this section we provide more comprehensive evaluation on the Argoverse~\cite{wilson2023argoverse} dataset in order to compare with the closely related work of Chang et al.~\cite{chang2023neural}.
We train our model on the 8 sequences selected by \cite{chang2023neural}, with their split of training and validation set.
We use their open-source code along with the released models to reproduce the rendering results of \cite{chang2023neural}.

We report the quantitative comparison in \cref{tab:argoversequantitative_supp}. It is evident that our method consistently outperforms \cite{chang2023neural} with a significant margin across all 8 sequences. In addition, we show more qualitative examples in \cref{fig:argoversequalitative_supp}. As can be seen, our rendering is not only complete (without blank pixels) but also with far higher resolution than the results from \cite{chang2023neural}.


\section{Additional Implementation Details}
\label{sec:details}
\subsection{Data Processing}

\noindent \textbf{Mask Dynamic Object.} 
We mask dynamic objects from images given annotated 2D bounding boxes. Specifically, a Mask-RCNN model takes as input the bounding boxes of dynamic objects, and outputs the corresponding masks. Pixel points within the dynamic object masks are not sampled during training and not counted when computing PNSR, SSIM and LPIPS. We also remove all Lidar points within the dynamic 3D bounding boxes. The remaining Lidar points are used to generate depth maps and augmented data. In PandaSet, the dynamic objects are in eight categories: 'Car', 'Pickup Truck', 'Medium-sized Truck', 'Semi-truck', 'Other Vehicle - Construction Vehicle', 'Other Vehicle - Uncommon', 'Other Vehicle - Pedicab', 'Emergency Vehicle' and 'Bus'.


\vspace{0.2cm}
\noindent \textbf{Lidar Depth Generation.}
In PandaSet, each Lidar frame is paired with a synchronized RGB frame. Lidar depth maps are generated from the accumulated training Lidar frames. Specifically, for a given image frame, LiDAR points from ten closest LiDAR frames are transformed into world coordinates and then projected onto the image coordinates. The depth value represents the distance along the ray from the camera's center to a LiDAR point. When multiple LiDAR points are projected onto the same pixel in the image, the depth value is derived from the nearest LiDAR point.

\vspace{0.2cm}
\noindent \textbf{Scene Normalization.} Scene normalization is required to apply the scene contraction strategy~\cite{barron2022mip} for handling unbounded scene. We do so based on the radius of camera trajectory. Specifically, we model the scene of interest with a sphere, whose diameter is the maximum distance between any two camera positions in the training log, plus 50 meters.

\subsection{Network Architectures}

Our network architecture can be decomposed into four main parts: proposal sampler, Lidar encoding, density network and color network.

\vspace{0.2cm}
\noindent \textbf{Proposal sampler} is composed of two consecutive neural density fields, each represented by ``fused" MLPs~\cite{tiny-cuda-nn}. These MLPs employ hash encoding to process 3D positions, followed by density prediction using an MLP. Both fused MLPs share common parameters: a minimum hash grid resolution of 16, five levels of hash encoding, 2-dim features for each hash encoding level, two MLP layers, and 16 hidden features in the MLP layers. The primary distinction between these two fused MLPs lies in their maximum hash grid resolution, with the first set at 512 and the second at 1024.

\vspace{0.2cm}
\noindent \textbf{Lidar encoding} takes as input all Lidar points and outputs high dimensional features for each Lidar point, which are then queried by sampled positions. To get started, the LiDAR points undergo voxelization within a grid of $512 \times 512 \times 512$ cells, with the cell features represented by the mean 3D positions of the LiDAR points they contain. Owing to the inherent sparsity, most cells remain empty, signifying an absence of LiDAR points and hence carrying negligible information. Subsequently, a 3D sparse UNet~\cite{choy20194d}, adhering to the encoder-decoder architecture similar to its 3D dense counterpart, is employed for encoding LiDAR geometry. This architecture integrates skip connections and multi-level feature fusion. Sparse convolution, a key component of this process, adapts the conventional convolution operation by applying filters exclusively to active (non-zero) input elements, thereby substantially enhancing computational efficiency and reducing memory demands. Within this architecture, the encoder's feature dimensions are set at 32, 64, 96, and 128, while the decoder's feature dimensions are 128, 96, 64, and 64. The Lidar features are the output of the final layer of this sparse UNet. Our implementation relies on the torchsparse~\cite{tangandyang2023torchsparse} package.

\vspace{0.2cm}
\noindent \textbf{Density network} computes hash encoding, then  fuses it with the Lidar encoding, and pass them to an MLP. Specifically, it is characterized by a hash grid with multiple resolutions ranging from a minimum of 16 to a maximum of 4096, and includes 16 levels of hash encoding, with each level comprising 2-dim features. The hash features queried from the hash grid are concatenated with the Lidar encoding and then passed to an MLP. 
The MLP component consists of two layers, each with 64-dim hidden features. In addition to predicting density values, the network also generates a 15-dimensional density embedding for each sampled position, which is subsequently fed into the color network.

\vspace{0.2cm}
\noindent \textbf{Color network} takes as input the density embedding and the ray direction encoded via spherical harmonics. It employs a two-layer MLP with 64-dim hidden features to predict an RGB value for each sampled position.

\subsection{Learning Hyper-parameters}

Our loss weights are set to $\lambda_1=0.0005$, $\lambda_2=1$, $\lambda_3=0.005$ and $\lambda_4=1$. We set all $\epsilon$ values in the unnormalized scale ($\epsilon^{0}_t=10m$, $\epsilon_t=100m$, $\epsilon^{0}_o=1m$, $\epsilon_o=0.15m$, $\epsilon_n=0.15m$, $\epsilon_a=1.5m$). The scheduling rate $\alpha_t$ and $\alpha_o$ is set to be 1.00004 and 0.99995, respectively. 

All modules in our network are optimized end-to-end for 100000 iterations, with RAdam~\cite{liu2019variance} algorithm. We utilize a per-iteration decay in the learning rate for each parameter. The learning rate is initially set at 0.01 and gradually reduced to 0.0001, with the learning rate scheduler having a maximum limit of 50,000 iterations. The number of sampled points per iteration is 4096.

%% file: figure/ds_ablation_supp.tex
\begin{figure*}[t!]
  \centering
  \includegraphics[width=1.0\linewidth]{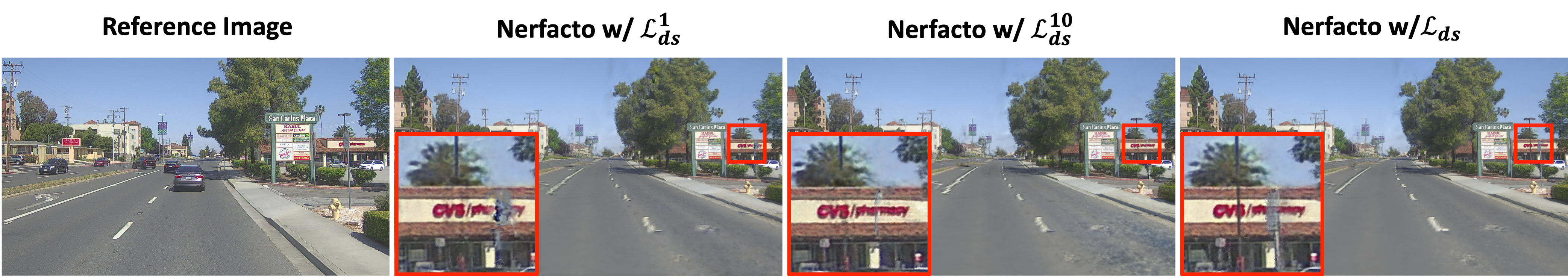}
  \centering
  \caption{
  \textbf{Qualitative comparison between different LiDAR depth supervisions on shift lane setting.} Our proposed method achieve significantly better rendering quality on the delicate structures.
  }
  \label{fig:ds_ablation_supp} 
  \vspace{-0.5em}
\end{figure*}

%% file: figure/aug_view_ablation_supp.tex
\begin{figure}
  \centering
  \includegraphics[width=1.0\linewidth]{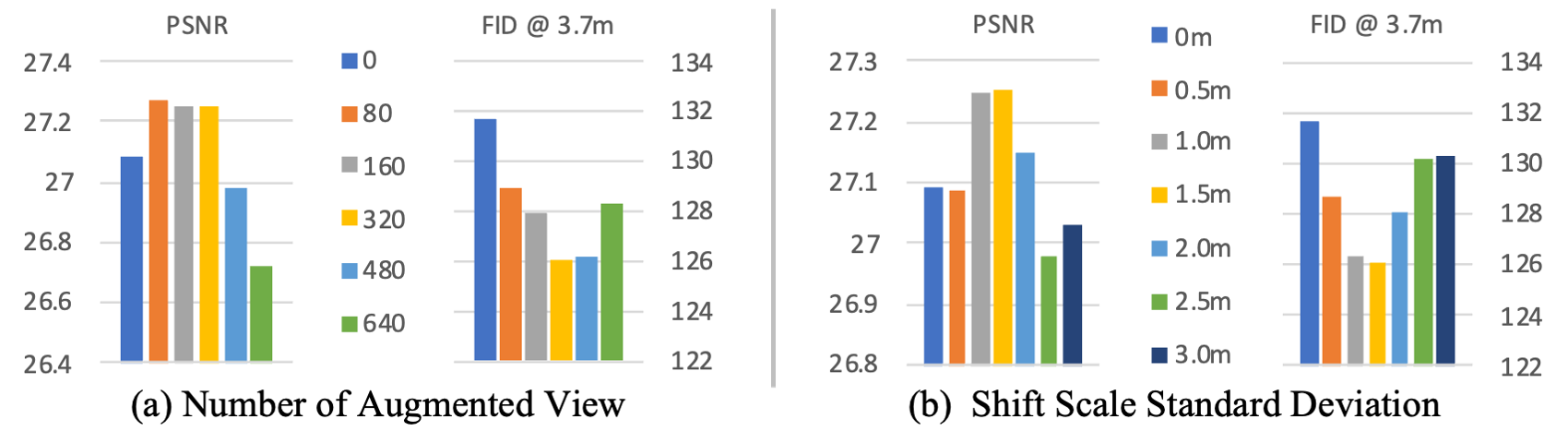}
  \centering
  \caption{
  Analysis on the impact of augmented synthetic views.}
  \label{fig:aug_ablation_supp} 
\end{figure}

%% file: tab/psnr_by_range.tex
\begin{table*}[htbp]
    \centering
    \resizebox{0.7\linewidth}{!}{
    \begin{tabular}{ccccccc}
    \toprule
    \multirow{2}{*}{Methods} & \multicolumn{3}{c}{Range by Distance} &
    \multicolumn{3}{c}{Range by Semantics} \\
    \cmidrule(r){2-4} \cmidrule(l){5-7}
    & Near ($\leq$ 10m) & Middle (10-60m) & Far ($\geq$60m) & Foreground & Boundary & Sky \\
    \midrule    
    UniSim & 25.44$|$25.92 & 25.55$|$26.07   & 22.41$|$21.99  & 25.44$|$26.32 & 22.07$|$22.23 & 34.81$|$36.61 \\
    Ours & {\textbf{26.68}$|$\textbf{27.62}} & \textbf{27.64}$|$\textbf{28.36} & \textbf{23.73}$|$\textbf{23.52} & \textbf{26.71}$|$\textbf{28.17} & \textbf{23.64}$|$\textbf{23.93} & \textbf{39.06}$|$\textbf{41.17} \\
    \bottomrule
    \end{tabular}}
    \caption{PSNR (mean$|$median) evaluation separated by range.}
    \label{tab:psnr_by_range}
\end{table*}

%% file: tab/lidar_encoding_ablation.tex
\begin{table}[]
    \centering
    \resizebox{1.0\linewidth}{!}{
    \begin{tabular}{lccccc}
    \toprule
    \multirow{2}{*}{Methods} & \multicolumn{3}{c}{Interpolation} &
    \multicolumn{2}{c}{Lane Shift} \\
    \cmidrule(r){2-4} \cmidrule(l){5-6}
    & {PSNR$\uparrow$ } & {SSIM$\uparrow$ } & {LPIPS$\downarrow$ } & {FID$\downarrow $ @ 2m} & {FID$\downarrow $ @ 3.7m} \\
    \midrule
    Original Hash   & 27.090 & 0.804 & 0.247 & 110.0 & 131.7 \\
    Double Hash     & 27.153 & 0.808 & 0.234 & 109.3 & 132.1 \\
    \hdashline
    MLP             & 27.119 & 0.805 & 0.246 & 108.0 & 131.6 \\
    PointNet++      & 27.076 & 0.804 & 0.247 & 108.7 & 131.2 \\
    \midrule
    Ours & \textbf{27.219} & \textbf{0.810} & \textbf{0.228} & \textbf{105.6} & \textbf{128.7} \\
    \bottomrule
    \end{tabular}}
	\caption{
            \textbf{Quantitative comparisons on Lidar encoding designs} -- Our proposed sparse convolution based Lidar encoding is better than simply doubling hash grid feature sizes and encoding Lidar feature with MLP or PointNet++ \cite{qi2017pointnet++}. ``Origina Hash'' indicates Nerfacto using our robust depth supervision. ``Double Hash'' doubles hash feature size.
        }
	\label{tab:lidar_encoding_ablation}
\vspace{-1em}
\end{table}

%% file: tab/argoverse.tex
\begin{table*}[]
    \centering
    \resizebox{1.0\linewidth}{!}{
    \begin{tabular}{ccccccccccccc}
    \toprule
    \multirow{2}{*}{Methods} & \multicolumn{3}{c}{Seq. 4d7b} &
    \multicolumn{3}{c}{Seq. 2b04} & \multicolumn{3}{c}{Seq. 4690} &
    \multicolumn{3}{c}{Seq. 0a13} \\
    \cmidrule{2-4} \cmidrule{5-7} \cmidrule{8-10} \cmidrule{11-13}
    & {PSNR$\uparrow$ } & {SSIM$\uparrow$ } & {LPIPS$\downarrow$ } & {PSNR$\uparrow$ } & {SSIM$\uparrow$ } & {LPIPS$\downarrow$ } & {PSNR$\uparrow$ } & {SSIM$\uparrow$ } & {LPIPS$\downarrow$ } & {PSNR$\uparrow$ } & {SSIM$\uparrow$ } & {LPIPS$\downarrow$ } \\
    \midrule
    Chang et al.~\cite{chang2023neural} & 24.319 & 0.652 & 0.127 & 22.257 &  0.598  & 0.123  & 26.049 & 0.728 & 0.181 & 20.318  &  0.615& 0.161 \\
    Ours         & \textbf{30.186} & \textbf{0.855} & \textbf{0.066} & \textbf{29.540} &  \textbf{0.906}  & \textbf{0.049} & \textbf{31.892} & \textbf{0.886}  & \textbf{0.077}  & \textbf{26.520} & \textbf{0.855}  & \textbf{0.110} \\
    \bottomrule    
    \bottomrule    
    \multirow{2}{*}{Methods} & \multicolumn{3}{c}{Seq. 2aea} &
    \multicolumn{3}{c}{Seq. 42c8} & \multicolumn{3}{c}{Seq. 4d32} &
    \multicolumn{3}{c}{Seq. 3e7c} \\
    \cmidrule{2-4} \cmidrule{5-7} \cmidrule{8-10} \cmidrule{11-13}
    & {PSNR$\uparrow$ } & {SSIM$\uparrow$ } & {LPIPS$\downarrow$ } & {PSNR$\uparrow$ } & {SSIM$\uparrow$ } & {LPIPS$\downarrow$ } & {PSNR$\uparrow$ } & {SSIM$\uparrow$ } & {LPIPS$\downarrow$ } & {PSNR$\uparrow$ } & {SSIM$\uparrow$ } & {LPIPS$\downarrow$ } \\
    \midrule
    Chang et al.~\cite{chang2023neural} & 24.901 & 0.706 & 0.098  & 22.990 & 0.708 & 0.161 & 25.186 & 0.706 & 0.143  &  26.440 & 0.720 & 0.143 \\
    Ours         & \textbf{31.996} & \textbf{0.885} & \textbf{0.071}  & \textbf{29.650} & \textbf{0.881} & \textbf{0.121} & \textbf{30.748} & \textbf{0.853} & \textbf{0.116}  & \textbf{33.876} & \textbf{0.912}  & \textbf{0.084} \\
    \bottomrule        
    \end{tabular}}
	\caption{
            \textbf{Quantitative comparisons on Argoverse with Chang et al.~\cite{chang2023neural}}.  We report PSNR, SSIM, and LPIPS on eight different sequences.
        }
	\label{tab:argoversequantitative_supp}
\vspace{-1em}
\end{table*}

%% file: tab/cdf_depth_loss.tex
\begin{table}[]
    \centering
    \resizebox{1.0\linewidth}{!}{
    \begin{tabular}{lccccc}
    \toprule
    \multirow{2}{*}{Methods} & \multicolumn{3}{c}{Interpolation} &
    \multicolumn{2}{c}{Lane Shift} \\
    \cmidrule(r){2-4} \cmidrule(l){5-6}
    & {PSNR$\uparrow$ } & {SSIM$\uparrow$ } & {LPIPS$\downarrow$ } & {FID$\downarrow $ @ 2m} & {FID$\downarrow $ @ 3.7m} \\
    \midrule
    Mid-Point & 26.945 & 0.795 & 0.270 & 112.7 & 139.0 \\
    CDF & \textbf{27.017} & \textbf{0.800} & \textbf{0.264} & \textbf{111.2} & \textbf{138.2} \\
    \bottomrule
    \end{tabular}}
	\caption{
            \textbf{Quantitative comparisons on depth loss implementation.} 
        }
	\label{tab:cdf_loss_ablation}
\vspace{-1em}
\end{table}

%% file: main.bbl
\begin{thebibliography}{52}
\providecommand{\natexlab}[1]{#1}
\providecommand{\url}[1]{\texttt{#1}}
\expandafter\ifx\csname urlstyle\endcsname\relax
  \providecommand{\doi}[1]{doi: #1}\else
  \providecommand{\doi}{doi: \begingroup \urlstyle{rm}\Url}\fi

\bibitem[git()]{githubGitHubLxxueFRNN}
{G}it{H}ub - lxxue/{F}{R}{N}{N}: {F}ixed {R}adius {N}earest {N}eighbor {S}earch on {G}{P}{U} --- github.com.
\newblock \url{https://github.com/lxxue/FRNN}.
\newblock [Accessed 18-11-2023].

\bibitem[wik()]{wikipediaInterstateHighway}
{I}nterstate {H}ighway standards - {W}ikipedia --- en.wikipedia.org.
\newblock \url{https://en.wikipedia.org/wiki/Interstate_Highway_standards}.
\newblock [Accessed 18-11-2023].

\bibitem[Aliev et~al.(2020)Aliev, Sevastopolsky, Kolos, Ulyanov, and Lempitsky]{aliev2020neural}
Kara-Ali Aliev, Artem Sevastopolsky, Maria Kolos, Dmitry Ulyanov, and Victor Lempitsky.
\newblock Neural point-based graphics.
\newblock In \emph{Computer Vision--ECCV 2020: 16th European Conference, Glasgow, UK, August 23--28, 2020, Proceedings, Part XXII 16}, pages 696--712. Springer, 2020.

\bibitem[Barron et~al.(2021)Barron, Mildenhall, Tancik, Hedman, Martin-Brualla, and Srinivasan]{barron2021mip}
Jonathan~T Barron, Ben Mildenhall, Matthew Tancik, Peter Hedman, Ricardo Martin-Brualla, and Pratul~P Srinivasan.
\newblock Mip-nerf: A multiscale representation for anti-aliasing neural radiance fields.
\newblock In \emph{Proceedings of the IEEE/CVF International Conference on Computer Vision}, pages 5855--5864, 2021.

\bibitem[Barron et~al.(2022)Barron, Mildenhall, Verbin, Srinivasan, and Hedman]{barron2022mip}
Jonathan~T Barron, Ben Mildenhall, Dor Verbin, Pratul~P Srinivasan, and Peter Hedman.
\newblock Mip-nerf 360: Unbounded anti-aliased neural radiance fields.
\newblock In \emph{Proceedings of the IEEE/CVF Conference on Computer Vision and Pattern Recognition}, 2022.

\bibitem[Caesar et~al.(2020)Caesar, Bankiti, Lang, Vora, Liong, Xu, Krishnan, Pan, Baldan, and Beijbom]{caesar2020nuscenes}
Holger Caesar, Varun Bankiti, Alex~H Lang, Sourabh Vora, Venice~Erin Liong, Qiang Xu, Anush Krishnan, Yu Pan, Giancarlo Baldan, and Oscar Beijbom.
\newblock nuscenes: A multimodal dataset for autonomous driving.
\newblock In \emph{Proceedings of the IEEE/CVF conference on computer vision and pattern recognition}, pages 11621--11631, 2020.

\bibitem[Carlson et~al.(2023)Carlson, Ramanagopal, Tseng, Johnson-Roberson, Vasudevan, and Skinner]{carlson2023cloner}
Alexandra Carlson, Manikandasriram~S Ramanagopal, Nathan Tseng, Matthew Johnson-Roberson, Ram Vasudevan, and Katherine~A Skinner.
\newblock Cloner: Camera-lidar fusion for occupancy grid-aided neural representations.
\newblock \emph{IEEE Robotics and Automation Letters}, 2023.

\bibitem[Chang et~al.(2023{\natexlab{a}})Chang, Chen, Ranjan, Yi, and Tuzel]{chang2023pointersect}
Jen-Hao~Rick Chang, Wei-Yu Chen, Anurag Ranjan, Kwang~Moo Yi, and Oncel Tuzel.
\newblock Pointersect: Neural rendering with cloud-ray intersection.
\newblock In \emph{Proceedings of the IEEE/CVF Conference on Computer Vision and Pattern Recognition}, pages 8359--8369, 2023{\natexlab{a}}.

\bibitem[Chang et~al.(2023{\natexlab{b}})Chang, Sharma, Kaess, and Lucey]{chang2023neural}
MingFang Chang, Akash Sharma, Michael Kaess, and Simon Lucey.
\newblock Neural radiance field with lidar maps.
\newblock In \emph{Proceedings of the IEEE/CVF International Conference on Computer Vision}, pages 17914--17923, 2023{\natexlab{b}}.

\bibitem[Choy et~al.(2019)Choy, Gwak, and Savarese]{choy20194d}
Christopher Choy, JunYoung Gwak, and Silvio Savarese.
\newblock 4d spatio-temporal convnets: Minkowski convolutional neural networks.
\newblock In \emph{Proceedings of the IEEE Conference on Computer Vision and Pattern Recognition}, pages 3075--3084, 2019.

\bibitem[Deng~et. al.(2022)]{kangle2021dsnerf}
Kangle Deng~et. al.
\newblock Depth-supervised {NeRF}: Fewer views and faster training for free.
\newblock In \emph{CVPR}, 2022.

\bibitem[Guo et~al.(2023)Guo, Deng, Li, Bai, Shi, Wang, Ding, Wang, and Li]{guo2023streetsurf}
Jianfei Guo, Nianchen Deng, Xinyang Li, Yeqi Bai, Botian Shi, Chiyu Wang, Chenjing Ding, Dongliang Wang, and Yikang Li.
\newblock Streetsurf: Extending multi-view implicit surface reconstruction to street views.
\newblock \emph{arXiv preprint arXiv:2306.04988}, 2023.

\bibitem[He et~al.(2017)He, Gkioxari, Doll{\'a}r, and Girshick]{he2017mask}
Kaiming He, Georgia Gkioxari, Piotr Doll{\'a}r, and Ross Girshick.
\newblock Mask r-cnn.
\newblock In \emph{Proceedings of the IEEE international conference on computer vision}, pages 2961--2969, 2017.

\bibitem[Hoetzlein(2014)]{hoetzlein2014fast}
Rama~C Hoetzlein.
\newblock Fast fixed-radius nearest neighbors: interactive million-particle fluids.
\newblock In \emph{GPU Technology Conference}, page~2, 2014.

\bibitem[Hu et~al.(2023{\natexlab{a}})Hu, Xu, Chu, and Jia]{hu2023trivol}
Tao Hu, Xiaogang Xu, Ruihang Chu, and Jiaya Jia.
\newblock Trivol: Point cloud rendering via triple volumes.
\newblock In \emph{Proceedings of the IEEE/CVF Conference on Computer Vision and Pattern Recognition}, pages 20732--20741, 2023{\natexlab{a}}.

\bibitem[Hu et~al.(2023{\natexlab{b}})Hu, Xu, Liu, and Jia]{hu2023point2pix}
Tao Hu, Xiaogang Xu, Shu Liu, and Jiaya Jia.
\newblock Point2pix: Photo-realistic point cloud rendering via neural radiance fields.
\newblock In \emph{Proceedings of the IEEE/CVF Conference on Computer Vision and Pattern Recognition}, pages 8349--8358, 2023{\natexlab{b}}.

\bibitem[Kajiya and Von~Herzen(1984)]{kajiya1984ray}
James~T Kajiya and Brian~P Von~Herzen.
\newblock Ray tracing volume densities.
\newblock \emph{ACM SIGGRAPH computer graphics}, 1984.

\bibitem[Katz et~al.(2007)Katz, Tal, and Basri]{katz2007direct}
Sagi Katz, Ayellet Tal, and Ronen Basri.
\newblock Direct visibility of point sets.
\newblock In \emph{ACM SIGGRAPH 2007 papers}, pages 24--es. 2007.

\bibitem[Kundu et~al.(2022)Kundu, Genova, Yin, Fathi, Pantofaru, Guibas, Tagliasacchi, Dellaert, and Funkhouser]{kundu2022panoptic}
Abhijit Kundu, Kyle Genova, Xiaoqi Yin, Alireza Fathi, Caroline Pantofaru, Leonidas~J Guibas, Andrea Tagliasacchi, Frank Dellaert, and Thomas Funkhouser.
\newblock Panoptic neural fields: A semantic object-aware neural scene representation.
\newblock In \emph{Proceedings of the IEEE/CVF Conference on Computer Vision and Pattern Recognition}, pages 12871--12881, 2022.

\bibitem[Liu et~al.(2023{\natexlab{a}})Liu, Chen, Yang, Wang, Manivasagam, and Urtasun]{liu2023real}
Jeffrey~Yunfan Liu, Yun Chen, Ze Yang, Jingkang Wang, Sivabalan Manivasagam, and Raquel Urtasun.
\newblock Real-time neural rasterization for large scenes.
\newblock In \emph{Proceedings of the IEEE/CVF International Conference on Computer Vision}, pages 8416--8427, 2023{\natexlab{a}}.

\bibitem[Liu et~al.(2019)Liu, Jiang, He, Chen, Liu, Gao, and Han]{liu2019variance}
Liyuan Liu, Haoming Jiang, Pengcheng He, Weizhu Chen, Xiaodong Liu, Jianfeng Gao, and Jiawei Han.
\newblock On the variance of the adaptive learning rate and beyond.
\newblock \emph{arXiv preprint arXiv:1908.03265}, 2019.

\bibitem[Liu et~al.(2023{\natexlab{b}})Liu, Tang, Amini, Yang, Mao, Rus, and Han]{liu2023bevfusion}
Zhijian Liu, Haotian Tang, Alexander Amini, Xinyu Yang, Huizi Mao, Daniela~L Rus, and Song Han.
\newblock Bevfusion: Multi-task multi-sensor fusion with unified bird's-eye view representation.
\newblock In \emph{2023 IEEE International Conference on Robotics and Automation (ICRA)}, 2023{\natexlab{b}}.

\bibitem[Mildenhall et~al.(2020)Mildenhall, Srinivasan, Tancik, Barron, Ramamoorthi, and Ng]{mildenhall2020nerf}
B Mildenhall, PP Srinivasan, M Tancik, JT Barron, R Ramamoorthi, and R Ng.
\newblock Nerf: Representing scenes as neural radiance fields for view synthesis.
\newblock In \emph{European conference on computer vision}, 2020.

\bibitem[M\"uller(2021)]{tiny-cuda-nn}
Thomas M\"uller.
\newblock {tiny-cuda-nn}, 2021.

\bibitem[M\"uller et~al.(2022)M\"uller, Evans, Schied, and Keller]{mueller2022instant}
Thomas M\"uller, Alex Evans, Christoph Schied, and Alexander Keller.
\newblock Instant neural graphics primitives with a multiresolution hash encoding.
\newblock \emph{ACM Trans. Graph.}, 2022.

\bibitem[Ost et~al.(2021)Ost, Mannan, Thuerey, Knodt, and Heide]{ost2021neural}
Julian Ost, Fahim Mannan, Nils Thuerey, Julian Knodt, and Felix Heide.
\newblock Neural scene graphs for dynamic scenes.
\newblock In \emph{Proceedings of the IEEE/CVF Conference on Computer Vision and Pattern Recognition}, pages 2856--2865, 2021.

\bibitem[Ost et~al.(2022)Ost, Laradji, Newell, Bahat, and Heide]{ost2022neural}
Julian Ost, Issam Laradji, Alejandro Newell, Yuval Bahat, and Felix Heide.
\newblock Neural point light fields.
\newblock In \emph{Proceedings of the IEEE/CVF Conference on Computer Vision and Pattern Recognition}, pages 18419--18429, 2022.

\bibitem[Qi et~al.(2017)Qi, Yi, Su, and Guibas]{qi2017pointnet++}
Charles~Ruizhongtai Qi, Li Yi, Hao Su, and Leonidas~J Guibas.
\newblock Pointnet++: Deep hierarchical feature learning on point sets in a metric space.
\newblock \emph{Advances in neural information processing systems}, 30, 2017.

\bibitem[Ravi et~al.(2020)Ravi, Reizenstein, Novotny, Gordon, Lo, Johnson, and Gkioxari]{ravi2020accelerating}
Nikhila Ravi, Jeremy Reizenstein, David Novotny, Taylor Gordon, Wan-Yen Lo, Justin Johnson, and Georgia Gkioxari.
\newblock Accelerating 3d deep learning with pytorch3d.
\newblock \emph{arXiv preprint arXiv:2007.08501}, 2020.

\bibitem[Rematas et~al.(2022)Rematas, Liu, Srinivasan, Barron, Tagliasacchi, Funkhouser, and Ferrari]{rematas2022urban}
Konstantinos Rematas, Andrew Liu, Pratul~P Srinivasan, Jonathan~T Barron, Andrea Tagliasacchi, Thomas Funkhouser, and Vittorio Ferrari.
\newblock Urban radiance fields.
\newblock In \emph{Proceedings of the IEEE/CVF Conference on Computer Vision and Pattern Recognition}, 2022.

\bibitem[Tancik et~al.(2022)Tancik, Casser, Yan, Pradhan, Mildenhall, Srinivasan, Barron, and Kretzschmar]{tancik2022block}
Matthew Tancik, Vincent Casser, Xinchen Yan, Sabeek Pradhan, Ben Mildenhall, Pratul~P Srinivasan, Jonathan~T Barron, and Henrik Kretzschmar.
\newblock Block-nerf: Scalable large scene neural view synthesis.
\newblock In \emph{Proceedings of the IEEE/CVF Conference on Computer Vision and Pattern Recognition}, pages 8248--8258, 2022.

\bibitem[Tancik et~al.(2023)Tancik, Weber, Ng, Li, Yi, Kerr, Wang, Kristoffersen, Austin, Salahi, Ahuja, McAllister, and Kanazawa]{nerfstudio}
Matthew Tancik, Ethan Weber, Evonne Ng, Ruilong Li, Brent Yi, Justin Kerr, Terrance Wang, Alexander Kristoffersen, Jake Austin, Kamyar Salahi, Abhik Ahuja, David McAllister, and Angjoo Kanazawa.
\newblock Nerfstudio: A modular framework for neural radiance field development.
\newblock In \emph{ACM SIGGRAPH 2023 Conference Proceedings}, 2023.

\bibitem[Tang et~al.(2023)Tang, Yang, Liu, Hong, Yu, Li, Dai, Wang, and Han]{tangandyang2023torchsparse}
Haotian Tang, Shang Yang, Zhijian Liu, Ke Hong, Zhongming Yu, Xiuyu Li, Guohao Dai, Yu Wang, and Song Han.
\newblock Torchsparse++: Efficient training and inference framework for sparse convolution on gpus.
\newblock In \emph{IEEE/ACM International Symposium on Microarchitecture (MICRO)}, 2023.

\bibitem[Vedaldi et~al.(2007)Vedaldi, Guidi, and Soatto]{vedaldi2007moving}
Andrea Vedaldi, Gregorio Guidi, and Stefano Soatto.
\newblock Moving forward in structure from motion.
\newblock In \emph{2007 IEEE Conference on Computer Vision and Pattern Recognition}, pages 1--7. IEEE, 2007.

\bibitem[Wang et~al.(2023{\natexlab{a}})Wang, Sun, Liu, Wu, Shen, Wu, Dai, and Zhang]{wang2023digging}
Chen Wang, Jiadai Sun, Lina Liu, Chenming Wu, Zhelun Shen, Dayan Wu, Yuchao Dai, and Liangjun Zhang.
\newblock Digging into depth priors for outdoor neural radiance fields.
\newblock In \emph{ACMMM}, 2023{\natexlab{a}}.

\bibitem[Wang et~al.(2023{\natexlab{b}})Wang, Louys, Piasco, Bennehar, Rold{\~a}o, and Tsishkou]{wang2023planerf}
Fusang Wang, Arnaud Louys, Nathan Piasco, Moussab Bennehar, Luis Rold{\~a}o, and Dzmitry Tsishkou.
\newblock Planerf: Svd unsupervised 3d plane regularization for nerf large-scale scene reconstruction.
\newblock \emph{arXiv preprint arXiv:2305.16914}, 2023{\natexlab{b}}.

\bibitem[Wang et~al.(2023{\natexlab{c}})Wang, Chen, Loy, and Liu]{wang2023sparsenerf}
Guangcong Wang, Zhaoxi Chen, Chen~Change Loy, and Ziwei Liu.
\newblock Sparsenerf: Distilling depth ranking for few-shot novel view synthesis.
\newblock In \emph{ICCV}, 2023{\natexlab{c}}.

\bibitem[Wang et~al.(2023{\natexlab{d}})Wang, Shen, Gao, Huang, Munkberg, Hasselgren, Gojcic, Chen, and Fidler]{wang2023neural}
Zian Wang, Tianchang Shen, Jun Gao, Shengyu Huang, Jacob Munkberg, Jon Hasselgren, Zan Gojcic, Wenzheng Chen, and Sanja Fidler.
\newblock Neural fields meet explicit geometric representations for inverse rendering of urban scenes.
\newblock In \emph{Proceedings of the IEEE/CVF Conference on Computer Vision and Pattern Recognition}, pages 8370--8380, 2023{\natexlab{d}}.

\bibitem[Wei et~al.(2023)Wei, Liu, Zhou, and Lu]{wei2023depth}
Yi Wei, Shaohui Liu, Jie Zhou, and Jiwen Lu.
\newblock Depth-guided optimization of neural radiance fields for indoor multi-view stereo.
\newblock \emph{PAMI}, 2023.

\bibitem[Wilson et~al.(2023)Wilson, Qi, Agarwal, Lambert, Singh, Khandelwal, Pan, Kumar, Hartnett, Pontes, et~al.]{wilson2023argoverse}
Benjamin Wilson, William Qi, Tanmay Agarwal, John Lambert, Jagjeet Singh, Siddhesh Khandelwal, Bowen Pan, Ratnesh Kumar, Andrew Hartnett, Jhony~Kaesemodel Pontes, et~al.
\newblock Argoverse 2: Next generation datasets for self-driving perception and forecasting.
\newblock \emph{arXiv preprint arXiv:2301.00493}, 2023.

\bibitem[Wimbauer et~al.(2023)Wimbauer, Yang, Rupprecht, and Cremers]{wimbauer2023behind}
Felix Wimbauer, Nan Yang, Christian Rupprecht, and Daniel Cremers.
\newblock Behind the scenes: Density fields for single view reconstruction.
\newblock In \emph{Proceedings of the IEEE/CVF Conference on Computer Vision and Pattern Recognition}, pages 9076--9086, 2023.

\bibitem[Xiao et~al.(2021)Xiao, Shao, Hao, Zhang, Chai, Jiao, Li, Wu, Sun, Jiang, et~al.]{xiao2021pandaset}
Pengchuan Xiao, Zhenlei Shao, Steven Hao, Zishuo Zhang, Xiaolin Chai, Judy Jiao, Zesong Li, Jian Wu, Kai Sun, Kun Jiang, et~al.
\newblock Pandaset: Advanced sensor suite dataset for autonomous driving.
\newblock In \emph{2021 IEEE International Intelligent Transportation Systems Conference (ITSC)}, pages 3095--3101. IEEE, 2021.

\bibitem[Xie et~al.(2022)Xie, Zhang, Li, Zhang, and Zhang]{xie2022s}
Ziyang Xie, Junge Zhang, Wenye Li, Feihu Zhang, and Li Zhang.
\newblock S-nerf: Neural radiance fields for street views.
\newblock In \emph{The Eleventh International Conference on Learning Representations}, 2022.

\bibitem[Xu et~al.(2022)Xu, Xu, Philip, Bi, Shu, Sunkavalli, and Neumann]{xu2022point}
Qiangeng Xu, Zexiang Xu, Julien Philip, Sai Bi, Zhixin Shu, Kalyan Sunkavalli, and Ulrich Neumann.
\newblock Point-nerf: Point-based neural radiance fields.
\newblock In \emph{Proceedings of the IEEE/CVF Conference on Computer Vision and Pattern Recognition}, pages 5438--5448, 2022.

\bibitem[Yang et~al.(2023{\natexlab{a}})Yang, Li, Zhou, Yuan, Liu, Yang, Qiu, and Shen]{yang2023nerfvs}
Chen Yang, Peihao Li, Zanwei Zhou, Shanxin Yuan, Bingbing Liu, Xiaokang Yang, Weichao Qiu, and Wei Shen.
\newblock Nerfvs: Neural radiance fields for free view synthesis via geometry scaffolds.
\newblock In \emph{Proceedings of the IEEE/CVF Conference on Computer Vision and Pattern Recognition}, pages 16549--16558, 2023{\natexlab{a}}.

\bibitem[Yang et~al.(2023{\natexlab{b}})Yang, Chen, Wang, Manivasagam, Ma, Yang, and Urtasun]{yang2023unisim}
Ze Yang, Yun Chen, Jingkang Wang, Sivabalan Manivasagam, Wei-Chiu Ma, Anqi~Joyce Yang, and Raquel Urtasun.
\newblock Unisim: A neural closed-loop sensor simulator.
\newblock In \emph{Proceedings of the IEEE/CVF Conference on Computer Vision and Pattern Recognition}, 2023{\natexlab{b}}.

\bibitem[Yin et~al.(2021)Yin, Zhou, and Krahenbuhl]{yin2021center}
Tianwei Yin, Xingyi Zhou, and Philipp Krahenbuhl.
\newblock Center-based 3d object detection and tracking.
\newblock In \emph{Proceedings of the IEEE/CVF conference on computer vision and pattern recognition}, 2021.

\bibitem[Yu et~al.(2022)Yu, Peng, Niemeyer, Sattler, and Geiger]{yu2022monosdf}
Zehao Yu, Songyou Peng, Michael Niemeyer, Torsten Sattler, and Andreas Geiger.
\newblock Monosdf: Exploring monocular geometric cues for neural implicit surface reconstruction.
\newblock \emph{NeurIPS}, 2022.

\bibitem[Zhang et~al.(2023{\natexlab{a}})Zhang, Kundu, Funkhouser, Guibas, Su, and Genova]{zhang2023nerflets}
Xiaoshuai Zhang, Abhijit Kundu, Thomas Funkhouser, Leonidas Guibas, Hao Su, and Kyle Genova.
\newblock Nerflets: Local radiance fields for efficient structure-aware 3d scene representation from 2d supervision.
\newblock In \emph{Proceedings of the IEEE/CVF Conference on Computer Vision and Pattern Recognition}, pages 8274--8284, 2023{\natexlab{a}}.

\bibitem[Zhang et~al.(2023{\natexlab{b}})Zhang, Huang, Ni, Li, and Zhang]{zhang2023frequency}
Yi Zhang, Xiaoyang Huang, Bingbing Ni, Teng Li, and Wenjun Zhang.
\newblock Frequency-modulated point cloud rendering with easy editing.
\newblock In \emph{Proceedings of the IEEE/CVF Conference on Computer Vision and Pattern Recognition}, pages 119--129, 2023{\natexlab{b}}.

\bibitem[Zhou et~al.(2018)Zhou, Park, and Koltun]{Zhou2018}
Qian-Yi Zhou, Jaesik Park, and Vladlen Koltun.
\newblock {Open3D}: {A} modern library for {3D} data processing.
\newblock \emph{arXiv:1801.09847}, 2018.

\bibitem[Zhou et~al.(2023)Zhou, Lin, Shan, Wang, Sun, and Yang]{zhou2023sampling}
Xiaoyu Zhou, Zhiwei Lin, Xiaojun Shan, Yongtao Wang, Deqing Sun, and Ming-Hsuan Yang.
\newblock Sampling: Scene-adaptive hierarchical multiplane images representation for novel view synthesis from a single image.
\newblock In \emph{ICCV}, 2023.

\end{thebibliography}
